%% file: arxiv.tex
\documentclass[11pt]{article}
\usepackage{arxiv}
\usepackage{graphicx} 
\usepackage{xcolor}
\usepackage{natbib}
\usepackage{amssymb}
\usepackage{amsmath}
\DeclareMathOperator*{\argmin}{arg\,min}
\usepackage{booktabs}
\usepackage{multirow}
\usepackage{cleveref}

\title{Towards Unified Attribution in Explainable AI, Data-Centric AI, and Mechanistic Interpretability}

\author{
    Shichang Zhang \\
    \texttt{shzhang@hbs.edu} \And
    Tessa Han \\
    \texttt{than@g.harvard.edu} \And
    Usha Bhalla \\
    \texttt{usha\_bhalla@g.harvard.edu} \And
    Himabindu Lakkaraju \\
    \texttt{hlakkaraju@hbs.edu}
}
\date{Harvard University}

\renewcommand{\shorttitle}{Unified Attribution}

\begin{document}

\maketitle

\begin{abstract}
\input{sections/abstract}
\end{abstract}

\section{Introduction}
\input{sections/introduction}

\section{The attribution problem}
\label{sec:problem}
\input{sections/problem_formulation}

\section{Unifying feature, data, and component attribution}
\label{sec:unifying}

\subsection{Understanding feature attribution (FA)}
\label{sec:feature}
\input{sections/feature_attribution}

\subsection{Understanding data attributions (DA)}
\label{sec:data}
\input{sections/data_attribution}

\subsection{Understanding component attributions (CA)}
\label{sec:component}
\input{sections/component_attribution}

\subsection{Common concepts among attribution methods}
\label{subsec:common_building_blocks}
\input{sections/common}

\section{Leveraging the unified perspective to advance research}
\label{sec:advance}
\input{sections/discussion}

\section{Alternative views}
\label{sec:alternative}
\input{sections/alternative}
\section{Conclusion}
\label{sec:conclusion}
\input{sections/conclusion}

\section{Acknowledgments}
\input{sections/acknowledgment}

\bibliographystyle{plainnat}
\bibliography{references/all_references}

\clearpage
\appendix
\section*{Appendix}
\input{sections/appendix}

\end{document}

%% file: sections/abstract.tex
The increasing complexity of AI systems has made understanding their behavior critical.
Numerous interpretability methods have been developed to attribute model behavior to three key aspects: input features, training data, and internal model components, which emerged from \textit{explainable AI}, \textit{data-centric AI}, and \textit{mechanistic interpretability}, respectively.
However, these attribution methods are studied and applied rather independently, resulting in a fragmented landscape of methods and terminology.
\textbf{This position paper argues that feature, data, and component attribution methods share fundamental similarities, and a unified view of them benefits both interpretability and broader AI research.}
To this end, we first analyze popular methods for these three types of attributions and present a unified view demonstrating that these seemingly distinct methods employ similar techniques (such as perturbations, gradients, and linear approximations) over different aspects and thus differ primarily in their perspectives rather than techniques.
Then, we demonstrate how this unified view enhances understanding of existing attribution methods, highlights shared concepts and evaluation criteria among these methods, and leads to new research directions both in interpretability research, by addressing common challenges and facilitating cross-attribution innovation, and in AI more broadly, with applications in model editing, steering, and regulation.


%% file: sections/introduction.tex
The escalating complexity of AI systems necessitates a deeper understanding of their behavior, giving rise to the field of AI \textit{interpretability}.
A central question in interpretability research is how to attribute an AI system's behavior to its underlying factors.
To address this question, researchers across different domains have developed various attribution methods.
\textit{Feature attribution (FA)}, which emerged from \textit{explainable AI}, quantifies the influence of input features on a model's output at test time to reveal critical decision drivers~\citep{zeiler2014visualizing, ribeiro2016should, horel2020significance, horel2022computationally, lundberg2017unified,smilkov2017smoothgrad}. \textit{Data attribution (DA)}, a core approach in \textit{data-centric AI}, analyzes how training data points shape overall model behavior during the training process~\citep{koh2017understanding, ghorbani2019data, ilyas2022datamodels}. \textit{Component attribution (CA)}, a key branch of \textit{mechanistic interpretability}, examines how internal model components, such as neurons or layers within a neural network (NN), contribute to model behavior~\citep{vig2020investigating, meng2022locating, nanda2023attribution, shah2024decomposing}.
Although numerous attribution methods and comprehensive survey papers have been published~\citep{guidotti2018survey, covert2021explaining, wang2024gradient, hammoudeh2024training, bereska2024mechanistic,rai2024practical, rauker2023toward, vilas2024position}, these methods have largely been studied and applied in isolation by their respective communities, resulting in a fragmented landscape of methods and terminology for similar concepts~\citep{saphra2024mechanistic}.


\textbf{We take the position that 1) feature, data, and component attribution share core techniques despite their different perspectives, and that 2) this more unified view advances both interpretability, by overcoming common challenges and stimulating cross-attribution innovation, and broader AI research, with applications in model editing, steering, and regulation.}
In the following sections, we first define the attribution problem and formalize a unified framework that encompasses all three types of attributions (\textsection \ref{sec:problem}).
Next, we examine each attribution type, analyzing its popular methods, and demonstrate how all three attribution types share fundamental techniques (i.e., perturbations, gradients, and linear approximations), key concepts, and evaluation criteria (\textsection \ref{sec:unifying}).
We summarize representative attribution methods based on their techniques in Table~\ref{tab:methods}.
Then, we demonstrate how this unified view can advance both interpretability and broader AI research (\textsection \ref{sec:advance}).
We also discuss alternative views of the paper's position (\textsection \ref{sec:alternative}).
In summary, we believe that this unified view enhances our understanding of attribution methods, bridges the current fragmented landscape, and provides new insights and research directions.



%% file: sections/problem_formulation.tex
Various attribution methods have been developed to analyze model behavior from different aspects.
In this section, we formally introduce three types of attribution problems and show how they fall under a unified framework.
Consider a learning problem with \( d \)-dimensional input features \( \mathbf{x} = [x_1, x_2, \ldots, x_d] \).
During training, a dataset of \( n \) data points:
\(
\mathcal{D}_{\text{train}} = \{\mathbf{x}^{(1)}, \mathbf{x}^{(2)}, \ldots, \mathbf{x}^{(n)}\}
\)
is used to train a model \( f_\theta \) with parameters \(\theta\) and components \( c = \{c_1, c_2, \ldots, c_m\} \) by optimizing the loss function \(\mathcal{L}(\theta)\).
At test (inference) time, the model generates an output \( f_\theta(\mathbf{x}^{\text{test}}) \) for a new input \( \mathbf{x}^{\text{test}} \).
For notational simplicity, we omit \(\theta\) and ``test" and use \( f \) and \( \mathbf{x} \) when the context is unambiguous.
A notation summary is provided in Appendix~\ref{app:notations}.
The core objective of all three problems is to attribute the model's output \( f(\mathbf{x}) \) to different aspects and quantify their influence with \textit{attribution scores}.

\textit{Feature attribution (FA)} quantifies how input features affect the model's output.
A features \( x_i \) may be pixels in images, tokens in text, or other domain-specific units with \( \phi_i(\mathbf{x}) \) as its attribution score.

\textit{Data attribution (DA)} analyzes how training data shape model behavior during training.
We quantify the influence of each training point \( \mathbf{x}^{(j)} \in \mathcal{D}_{\text{train}} \) through its attribution score \( \psi_j(\mathbf{x}) \).

\textit{Component attribution (CA)} studies the role of model components in producing model outputs.
The component \( c_k \) can have various definitions, like a neuron in an NN, with \( \gamma_{k}(\mathbf{x}) \) as its attribution score.

\begin{figure*}[t]
    \centering
    \includegraphics[width=\textwidth]{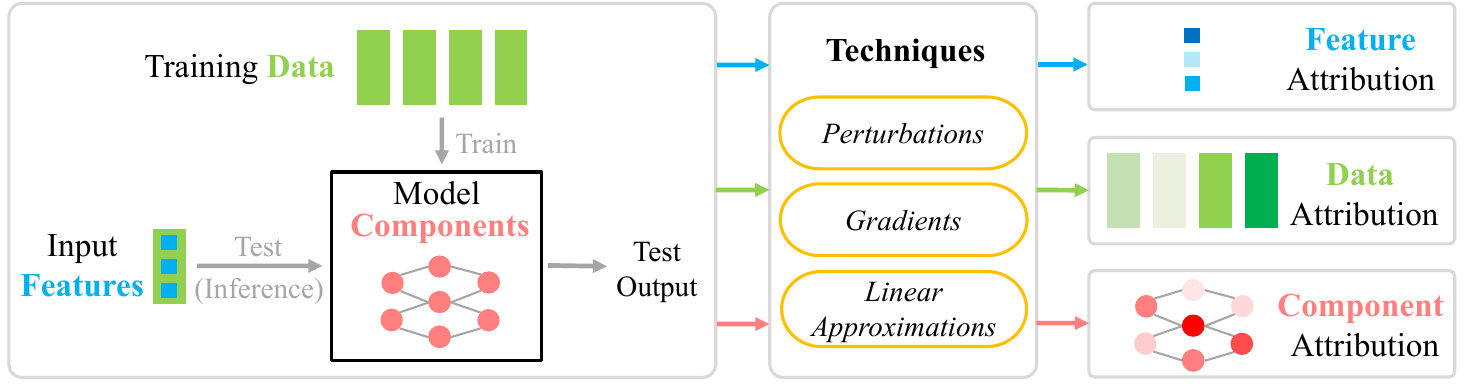}
    \caption{The three types of attribution: FA, DA, and CA. While each type seeks to attribute a model's output to a different aspect (input features, training data, and model components) and provides complementary insight into model behavior, they all use shared techniques (perturbations, gradients, and linear approximations).}
    \label{fig:attribution_framework}
\end{figure*}

As illustrated in Figure~\ref{fig:attribution_framework}, these three attribution problems share a fundamental connection: they all seek an \textit{attribution function} \( g \) that assigns scores to specific aspects (features \( x_i \), training points \( \mathbf{x}^{(j)} \), or components \( c_k \)) for a given test output \( f(\mathbf{x}) \), differing only in the choice of aspects.

\input{tables/methods}

%% file: tables/methods.tex
\begin{table}[tb]
    \centering
    \caption{Representative FA, DA, and CA methods classified by core techniques into three categories showing a unified view that these three types differ mainly in perspective but share techniques.}
    \resizebox{\textwidth}{!}{%
    \begin{tabular}{lllll}
    \toprule
    \textbf{Technique} & \textbf{Technique Details} & \textbf{Feature Attribution} & \textbf{Data Attribution} & \textbf{Component Attribution} \\ \midrule
    \multirow{1}{*}{\textbf{Perturbation}} & 
    \multirow{1}{*}{Direct} & Occlusions~\citep{zeiler2014visualizing} & LOO~\citep{cook1982residuals} & Causal Tracing~\citep{meng2022locating} \\
    & & RISE~\citep{petsiuk2018rise} & & Path Patching~\citep{wang2022interpretability} \\
    & & & & \citet{vig2020investigating}\\
    & & & & \citet{bau2020understanding}\\
    & & & & ACDC~\citep{conmy2023towards}\\
    \cmidrule{2-5}
    & \multirow{1}{*}{Game-Theoretic} & SHAP~\citep{lundberg2017unified} & Data Shapley~\citep{ghorbani2019data} & Neuron Shapley~\citep{ghorbani2020neuron} \\ 
    & \multirow{1}{*}{(Shapley)} & & TMC Shapley~\citep{ghorbani2019data} & \\ 
    & & & KNN Shapley~\citep{jia2019efficient} & \\ 
    & & & Beta Shapley~\citep{kwon2022beta} & \\ \cmidrule{2-5}
    & \multirow{1}{*}{Game-Theoretic} & STII~\citep{dhamdhere2019shapley} & Data Banzhaf~\citep{wang2023data} & -- \\ 
    & (Others)& BII~\citep{patel2021high} & & \\ 
    & & Core Value~\citep{yan2021if} & & \\ 
    & & Myerson Value~\citep{chen2018shapley} & & \\
    & & HN Value~\citep{zhang2022gstarx} &  & \\ \cmidrule{2-5}
    & \multirow{1}{*}{Mask Learning} & \citet{dabkowski2017real} & -- & \citet{csordas2020neural} \\
    & & L2X~\citep{pmlr-v80-chen18j} &  & Subnetwork Pruning~\citep{cao2021low} \\
    \midrule
    \multirow{1}{*}{\textbf{Gradient}} 
    & \multirow{1}{*}{First-Order} & Vanilla Gradients~\citep{simonyan2013deep} & GradDot/GradCos~\citep{pruthi2020estimating} & Attribution Patching~\citep{nanda2023attribution} \\
    & & Gradient $\times$ Input~\citep{shrikumar2017learning} & & EAP~\citep{syed2023attribution} \\
    & & SmoothGrad~\citep{smilkov2017smoothgrad} &  & \\
    & & GBP~\citep{springenberg2014striving} & & \\
    & & Grad-CAM~\citep{selvaraju2016grad} & & \\ \cmidrule{2-5}
    & \multirow{1}{*}{Second-Order} & Integrated Hessian~\citep{janizek2021explaining} & IF~\citep{koh2017understanding} & -- \\
    & (Hessian/IF) & & FastIF~\citep{guo-etal-2021-fastif} & \\
    & & & Arnoldi IF~\citep{schioppa2022scaling} & \\
    & & & EK-FAC~\citep{grosse2023studying} & \\
    & & & RelateIF~\citep{barshan2020relatif} & \\ \cmidrule{2-5}
    & \multirow{1}{*}{Tracing Path} & Integrated Grad~\citep{sundararajan2017axiomatic} & TracIn~\citep{pruthi2020estimating} & Attribution Path Patching~\citep{nanda2023attribution} \\ 
    & &  & SGD-Influence~\citep{hara2019data} & \\ 
    & &  & SOURCE~\citep{bae2024training} & \\ 
    \midrule
    \multirow{1}{*}{\textbf{Linear Approximation}} & 
    \multirow{1}{*}{} & LIME~\citep{ribeiro2016should} & Datamodels~\citep{ilyas2022datamodels} & COAR~\citep{shah2024decomposing} \\ 
    & & C-LIME~\citep{agarwal2021towards} & TRAK~\citep{park2023trak} & \\ \bottomrule
    \end{tabular}%
    }
    \label{tab:methods}
    \end{table}

%% file: sections/feature_attribution.tex
FA quantifies how individual features \(x_i\) of an input \(\mathbf{x}\) influence the model's output \(f(\mathbf{x})\) through attribution scores \(\phi_i(\mathbf{x})\).
Applied to model inference at test time, it explains model behavior without altering model parameters.
The attribution results can help with feature selection, model debugging, and improve user trust.
FA methods can be broadly classified into three categories: \textit{perturbation-based}, \textit{gradient-based}, and \textit{linear approximation}.
We discuss some prominent methods in each category below and provide more details in \Cref{app:fa}.

\textbf{Perturbation-Based FA.}
These methods attribute feature importance by measuring how model outputs change when input features are modified and especially removed (also called \textit{removal-based } methods~\citep{covert2021explaining}).
\textit{Direct perturbation}
straightforwardly applies this idea, e.g.,
the pioneering Occlusion method~\citep{zeiler2014visualizing} in computer vision (CV) replaces image pixels with grey squares and measures prediction changes.
For images, pixel attribution scores create a \textit{saliency map} highlighting the most influential regions.
RISE~\citep{petsiuk2018rise} advances this method by perturbing multiple image regions and combining the results.
While intuitive, direct perturbation fails to capture synergistic interactions between multiple features. \textit{Game-theoretic perturbation} addresses this limitation by modeling features as game players collaborating toward the model's output.
The Shapley value~\citep{shapley1953value} provides a foundational solution within this framework and has inspired numerous FA methods~\citep{sundararajan2020many}.
Computing Shapley value attributions involves measuring a specific type of perturbation: how adding a 
feature \(x_i\) to different feature subsets changes the model's output compared to the subset alone, 
known as the marginal contribution of \(x_i\) to the subset.
The final attribution score captures feature interactions by aggregating these marginal contributions across all possible feature subsets, which can be computationally challenging.
Various approximation methods have been proposed for this computation, and 
KernelSHAP \cite{lundberg2017unified} (or simply SHAP) has gained widespread 
adoption because of its efficient kernel-based approximation.
Since removing input features can be viewed as applying binary masks for features, \textit{perturbation mask learning} methods use learnable masks representing feature removal probabilities, which offer more nuanced control compared to binary masks.
A new mask can be learned for each input~\citep{fong2017interpretable} or only once as a global \textit{masking model}~\citep{dabkowski2017real}.
The masking model acts as the attribution function \(g\). While initial training is required, \(g\) can be faster at test time by generating masks through a single forward pass.

\textbf{Gradient-based FA.} 
Gradients of model outputs \(f(\mathbf{x})\) with respect to input features \(\mathbf{x}\), \(\nabla_\mathbf{x} f(\mathbf{x})\), quantify output sensitivity to small input changes~\citep{erhan2009visualizing, baehrens2010explain}, measuring feature influence without requiring perturbations.
Gradient-based methods are more computationally efficient, needing only a single or a few forward and backward pass(es) to compute \(\nabla_\mathbf{x} f(\mathbf{x})\) instead of at least \(O(d)\) model perturbations for \(d\) features.
Also originating from CV, these methods gained widespread adoption for generating image saliency maps~\citep{simonyan2013deep} (also called \textit{sensitivity maps}~\citep{smilkov2017smoothgrad}).
The ``vanilla gradients'' method uses the gradients of the output class (log)probability with respect to input pixels as attribution scores~\citep{simonyan2013deep}.
Since then, numerous enhanced gradient-based methods has been proposed.
For example, 
Integrated Gradients~\citep{sundararajan2017axiomatic} accumulates gradients along a path from a baseline to the actual input, and Integrated Hessians~\citep{janizek2021explaining} further extends the analysis to feature interactions by computing the Hessian matrix.
SmoothGrad~\citep{smilkov2017smoothgrad} generates multiple copies of the input with added Gaussian noise and computes sensitivity maps for each noisy sample.
It then averages these maps to reduce noise while preserving salient features for model outputs.

\textbf{Linear approximation for FA.}
These methods offer an alternative approach to FA by approximating the complex behavior of \(f\) around an input of interest \(\mathbf{x}\) using a linear model, typically in the form of \(g(\mathbf{x}) = \mathbf{w}^\top \mathbf{x} + b\) with coefficients \(\mathbf{w}\) and bias \(b\), and using coefficient \(w_i\) as the FA score of feature \(x_i\).
LIME~\citep{ribeiro2016should} exemplifies this approach.
It fits a sparse linear model to capture local model behavior and uses binary indicators (0 or 1) rather than actual feature values as inputs to the linear model, only representing feature exclusion or inclusion.
The resulting linear model coefficients explain how each feature's presence influences the approximated model's output. 
Notably, LIME can also be viewed through a perturbation lens, as it fundamentally perturbs input features to approximate the output.
This connection suggests that different FA methods can be understood within a common mathematical framework, which we explore next.

\subsubsection{Unifying feature attribution methods via local function approximation (LFA)}
\label{sec:lfa}

While FA methods can be grouped in the three categories above, many of them can be unified under the LFA framework~\citep{lfa2022}.
In this framework, a model \(f\) is approximated around a point of interest \(x\) in a local neighborhood \(\mathcal{Z}\) by an interpretable model \(g\) using a loss function \(\ell\).
Eight prominent FA methods (Occlusion, KernelSHAP, Vanilla Gradients, Gradients $\times$ Input, Integrated Gradients, SmoothGrad, LIME, and C-LIME) are instances of this framework, differing only in their choices of local neighborhoods \(\mathcal{Z}\) and loss functions \(\ell\) (Appendix Table~\ref{tab:lfa})~\cite{lfa2022}.

The LFA framework~\citep{lfa2022} enhances our understanding of FA methods in several ways.
First, it provides conceptual coherence to the field.
While different methods appear to have distinct motivations, this framework reveals their shared goal of LFA.
Second, placing diverse methods under a single framework enables direct comparisons among them.
This comparative lens allows us to better understand their similarities, differences, and behavior, such as why different methods can generate disagreeing explanations for the same model prediction \citep{disagreement2024}.
Third, this unification enables theoretical simplicity. Instead of studying methods individually, theoretical analyses can be performed using the framework and applied to each method~\cite{lfa2022, bilodeau2024impossibility, bordt2023shapley}.
Fourth, the conceptual understanding brought about by unification leads to principled, practical recommendations \citep{lfa2022}. Additional details on this unification are discussed in Appendix~\ref{app:unify-fa}.

%% file: sections/data_attribution.tex
DA studies how the training dataset \(\mathcal{D}_{\text{train}}\) shapes model behavior during training.
It is also known as \textit{data valuation}, as it assesses the value of data from vendors.
For each training data point \(\mathbf{x}^{(j)}\), an attribution score \(\psi_j(\mathbf{x})\) traces back to training to quantify \(\mathbf{x}^{(j)}\)'s influence on the model's output \(f(\mathbf{x})\) for a test point \(\mathbf{x}\).
DA scores can be used to characterize data properties, help identify mislabeled data, and justify training data values.
Like FA, DA uses the techniques of perturbations, gradients, and linear approximation.
We examine prominent methods below, with additional details in~\Cref{app:da}.

\textbf{Perturbation-based DA}. 
These DA methods examine changes in the model behavior after removing or re-weighting the training data points and subsequently retraining the model (as a result, they are also called \textit{retraining-based methods}~\citep{hammoudeh2024training}).
Leave-One-Out (LOO) attribution, analogous to direct perturbation in FA, is a prominent example that trains a model on the complete training set and then separately removes each individual data point and retrains the model.
The attribution score for each removed point is determined by the corresponding performance difference.
The LOO approach has a long history in statistics~\citep{cook1982residuals} and has proven useful for DA for modern AI models~\citep{jia2021scalability}. 
It provides valuable insights with its main limitation being the computational cost of model retraining for each data point.
Many DA methods later can be viewed as efficient approximations of LOO.
A natural extension of LOO is to leave a set of data points out to evaluate their collective impact~\citep{ilyas2022datamodels}.
As the direct perturbation for FA, LOO overlooks interactions between data points, potentially missing subtle influence behaviors~\citep{lin2022measuring, jia2021scalability}.
This is addressed by game-theoretic DA methods like Data Shapley~\citep{ghorbani2019data} which applies the Shapley value to DA by computing each training point's aggregated marginal contribution across all training subsets.
Similar to game-theoretic FA, they face prohibitive computational costs, as each marginal contribution requires retraining and there are \(2^n\) possible subsets.
Many approximations have been proposed to address this challenge (\Cref{app:da_game}).

\textbf{Gradient-based DA}.
These DA methods use the gradients of the loss with respect to both training data \(\nabla_\theta \mathcal{L}(f_\theta(\mathbf{x}^{(j)}))\) and test data \(\nabla_\theta \mathcal{L}(f_\theta(\mathbf{x}))\) to assess the impact of training point \(\mathbf{x}^{(j)}\) on model output \(f(\mathbf{x})\).
The GradDot and GradCos methods use the dot product and cosine similarity of these two gradients as attribution scores \(\psi_j(\mathbf{x})\)~\cite{charpiat2019input}.
Influence Function (IF), a classic statistical method for analyzing influential points in linear regression~\citep{cook1980characterizations}, has been adapted for modern AI models~\citep{koh2017understanding}.
IF approximates LOO model parameter changes by Taylor expansion, computing both the gradient and the (inverse) Hessian, thus avoiding retraining.
It offers an effective and computationally feasible alternative to LOO but also faces challenges including convexity assumptions and Hessian matrix computation which many follow-up methods try to address (\Cref{app:da_if}).
Aside from these IF-based methods, TracIn~\citep{pruthi2020estimating} attributes the data influence throughout the entire training path by computing dot products between training and test data gradients at each training step from the initial model parameters to the final model parameters, accumulating these to capture a total influence across the training path.
This path tracing method avoids limitations of LOO and IF, such as assigning identical attribution scores to duplicate training data points.

\textbf{Linear approximation for DA.}
Datamodel~\citep{ilyas2022datamodels} applies linear approximation to DA, similar to LIME in FA.
It constructs a linear model \(g\) with \(n\) coefficients and \(\{0,1\}^n\) vectors as inputs, where each input represents a subset of training data. \(g\) is learned to map any subset of training data to output \(f(\mathbf{x})\), where \(f\) is trained on this subset with the given model architecture and training algorithm.
The coefficients of \(g\) thus represent the attribution scores of the data points.
While Datamodel can effectively capture model behavior, constructing this large linear model with \(n\) coefficients requires extensive counterfactual data obtained by training model \(f\) on various subsets, making it computationally intensive.
TRAK~\citep{park2023trak} addresses the computational challenge by estimating  Datamodels in a transformed space where the learning problem becomes convex.
Furthermore, both approaches can be viewed as perturbation-based methods, similar to LIME, as they vary training data to construct linear models.


%% file: sections/component_attribution.tex
CA studies the role of internal model components in model outputs.
It plays a central role in the emerging \textit{mechanistic interpretability} research which seeks to understand AI models by reverse engineering their internal mechanisms into interpretable algorithms.
CA operates primarily at test time for model inference, quantifying how each model component \(c_k\) contributes to the output \(f(\mathbf{x})\) through an attribution score \(\gamma_{k}(\mathbf{x})\).
Components can be defined flexibly - from individual neurons and attention heads to entire layers and \textit{circuits} (subnetworks).
Like FA and DA, CA leverages the same three techniques: perturbations, gradients, and linear approximation.
We examine key CA methods below and provide additional details in Appendix~\ref{app:ca}.

\textbf{Perturbation-based CA.}
Perturbation-based methods for CA are fundamentally similar to those for FA and DA.
Components of the model, whether neurons, layers, or circuits, are perturbed to measure their effect on model behavior. Generally, the perturbations are chosen carefully to attempt to localize behaviors related to specific tasks or concepts. 
\textit{Causal mediation analysis}~\citep{pearl2022direct, vig2020investigating} is a perturbation-based method that leverages the abstraction of models to causal graphs, with graphs nodes corresponding to components, and graph edges represent the causal relationships between nodes. 
Causal mediation analysis is defined by an input cause \(\mathbf{x}\) and an output effect \(f(\mathbf{x})\) that is mediated by intermediate causal nodes, \(c_k\), between \(\mathbf{x}\) and \(f(\mathbf{x})\). By perturbing \(c_k\), changes in \(f(\mathbf{x})\) can be measured to get attribution scores \(\gamma_{k}(\mathbf{x})\). 
Specifically, the effects of perturbations on component activations are measured during three separate runs: a clean run with no perturbations, a corrupted run where components are perturbed (may also be perturbing inputs to induce corrupted component activations), and a corrupted-with-restoration run that measures how much restoring a single component activation of the corrupted run by its original clean-run value can restore the output. 
The corrupted run can be repeated multiple times with different random noise added to obtain more robust CA scores.
This analysis is frequently referred to as \textit{causal tracing}~\citep{meng2022locating} or \textit{activation patching}, and \textit{path patching}~\citep{wang2022interpretability} when patching is applied to paths connecting components. 
\textit{Game-theoretic CA}~\citep{ghorbani2020neuron} follows a similar approach to game-theoretic FA and DA. 
In particular, Neuron Shapley~\citep{ghorbani2020neuron} extends prior works on Shapley values to CA and mitigates computational cost through sampling and a multi-armed bandit algorithm to identifies neurons with large attribution scores. 
\textit{Mask Learning and Subnetwork Probing}~\citep{csordas2020neural, cao2021low} adopts a similar concept to FA mask learning, attempting to approximate either the model's or a probe's performance on a given task by searching for components that equivalently perform that task. More specifically, subnetwork probing optimizes a mask for the model weights as a continuous relaxation for searching, which essentially prunes the model.


\textbf{Gradient-based CA.}
To improve the computational efficiency of CA methods, alterations of causal tracing leverage gradient-based approximations to do only two forward passes and a single backward pass to generate attributions.
Attribution patching~\citep{nanda2023attribution} approximates the local change when patching a single activation from the corrupted to clean input.
This is achieved by computing the backward pass for the corrupted output with respect to the patching metric and storing the gradients with respect to the activations. 
Finally, the difference between the clean and corrupted activations is taken and multiplied by the cached gradients to obtain attribution scores.

\textbf{Linear approximation for CA.}
CA also employs linear approximations like LIME in FA and Datamodels in DA. COAR~\citep{shah2024decomposing} attempts to decompose model behavior in terms of various components by predicting the counterfactual impact of ablating each component. Given the computational complexity, COAR employs linear approximations by assigning scores to each component and estimating the counterfactual effect of removing sets of components by summing their scores. Thus, the complexity of relationships between components is abstracted away. 

%% file: sections/common.tex
As demonstrated in the previous sections, FA, DA, and CA methods leverage three shared techniques: perturbations, gradients, and linear approximations. 
Additional details are discussed in \Cref{app:common_methods} and a summary of the methods is provided in \Cref{tab:methods}.
Beyond shared techniques, conceptual ideas also transfer across the three types of attributions.
One such idea is the deliberate introduction of randomness and smoothing to enhance attribution robustness.
This idea has proven effective across all three attribution types, with SmoothGrad using random noisy samples to denoise gradients for FA, TRAK ensembling attributions over multiple retrainings for DA, and causal mediation analysis aggregating results from multiple corrupted runs for CA.
Another shared conceptual idea is tracking and aggregating results along paths. For example, Integrated Gradients calculates the path integral from base input to target input in FA, TracIn traces training paths to reveal dynamic data influences in DA, and path patching tracks component effects along residual stream paths in CA. 
These shared concepts are high-level ideas, not specific algorithms, and the definitions of paths can be different, but they highlight fundamental connections among different types of attributions.
For example, both Integrated Gradients and TracIn admit inspiration from the Fundamental Theorem of Calculus.

Common approaches have also been utilized to evaluate the quality of all three types of attribution~~\citep{agarwal2022openxai,hammoudeh2024training,rai2024practical}.
Counterfactual evaluation is a widely used approach, which compares the attribution scores with the actual impact of removing or modifying inputs. For example, evaluate by measuring the model's performance drop when the inputs with high attribution scores are removed.
Task-specific evaluation is another common approach to measure the attribution quality by running specific downstream tasks with the attribution results. In this type of evaluation, one can use FA to help identify feature changes that can flip model outputs, apply DA to detect mislabeled training examples, and run CA to identify important components that allows for model pruning.
Moreover, human evaluation in many cases can be a gold standard by relying on domain experts or users to assess the quality and interpretability of attributions.
More details on these evaluation approaches and their pros and cons are provided in \Cref{app:dis_eval}.

%% file: sections/discussion.tex
Feature, data, and component attribution have been studied largely as separate research areas, resulting in the parallel development of similar methods from different communities with distinct terminologies.
Having demonstrated the unified view of the three types of attribution, we now discuss how this unified view can help advance both interpretability and broader AI research.
We point out open challenges as well as potential solutions as future research directions.

\subsection{Advancing research in interpretability}
\label{subsec:adv_attribution_research}

\subsubsection{Overcoming shared challenges among all attribution types}
\label{subsec:common_challenges}

Attribution methods face several common challenges that need to be addressed to enhance their practicality and reliability.
Our unified perspective provides clearer insights into the underlying causes of these challenges and their potential solutions.
We discuss these key challenges and propose potential solutions both below and in greater detail in \cref{app:common_challenges}.

\textbf{Improving computational efficiency.} 
Computational efficiency is a significant challenge for all three types of attribution methods, preventing them from being applied to large models~\citep{adolfi2024computational, bassan2024local, barcelo2020model}.
Our unified view reveals that this challenge is rooted in their shared techniques.
For perturbation-based methods, the curse of dimensionality makes it intractable to comprehensively analyze high-dimensional inputs, large training datasets, and models with numerous components.
Gradient-based methods offer more practical computational costs, except when sophisticated gradient computations are needed, such as aggregating gradients or computing second-order Hessian matrices.
Linear approximation methods also face computational challenges when numerous data points and model evaluations are required to establish sufficient data for learning accurate linear models.
Therefore, improving computational efficiency calls for more efficient sampling methods (e.g., Maximum Sample Reuse~\citep{wang2023data}) to avoid perturbing the entire input space in perturbation-based methods and to learn more accurate linear models for linear approximation methods. Also needed are more efficient gradient computations, especially for higher-order gradients like Hessians. Many works have sought to improve these aspects, with promising examples like \citet{grosse2023studying} showing that gradient-based methods can be scaled to large language models (LLMs) with up to 52 billion parameters. However, there is still much room for improvement to make attribution methods more efficient.
Also, a unified perspective may be developed for the corresponding computational problems~\citep{adolfi2024computational, bassan2024local, barcelo2020model}.

\textbf{Improving attribution consistency.} 
Consistency challenges refer to the variability in attribution results across multiple runs of the same method with different random seeds, making it challenging to establish stable interpretations and evaluations~\citep{jeyakumar2020can}.
Our unified view reveals that this challenge is prevalent across all three attribution types due to multiple of randomness rooted in their common techniques, including sampling, learning processes with stochastic optimization, and non-trivial hyperparameters.
We see the potential improvement for this challenge including producing more deterministic attribution methods that are less sensitive to random hyperparameters, provide provable guarantees of attribution stability or error bounds. While past works have made some progress in this direction, there were strong assumptions added to the setting, e.g., convexity~\citep{koh2017understanding}. Thus, more efforts are needed to improve consistency in more realistic settings.

\textbf{Towards fairer and more practical evaluation.} 
Evaluation challenges arise in all three types of attribution due to multiple factors.
Inconsistent results make it difficult to reliably compare methods and determine their relative accuracy.
This challenge is particularly evident in gradient-based FA methods.
As demonstrated by \citet{adebayo2018sanity}, they can produce contradictory attribution results and sometimes perform no better than random baselines, making them difficult to evaluate fairly.
For the shared evaluation metrics among all three types of attribution discussed in \Cref{subsec:common_building_blocks}, counterfactual evaluation could provide more rigorous validation, but computational constraints often make this approach impractical.
Task-specific evaluations offer easier alternatives but frequently lack generalizability across different contexts.
Human evaluation, despite being considered the gold standard, faces scalability issues and potential biases.
The diversity of evaluation metrics and their varying definitions of importance make the evaluation more challenging.
An attribution result may perform well by one metric but poorly according to another.
These challenges emphasize the pressing need for developing more reliable and generalizable attribution evaluation metrics.

Altogether, these are open challenges that call for future research to develop more efficient and reliable attribution methods as well as rigorous and principled frameworks to evaluate them. Given the shared techniques and common challenges among different attribution types, ideas to address these challenges in one attribution type will likely also be relevant to others.

\subsubsection{Cross-attribution innovation}
\label{subsec:cross_aspect_inspiration}

The connections among FA, DA, and CA discussed in the sections above naturally suggest promising research directions of leveraging insights from one type of attribution to develop methods for another. 
This can be directly identified as filling in the empty cells in Table~\ref{tab:methods}.
For example, while the Shapley value has been successfully applied across all three types of attribution, many other game-theoretic notions have only been used in FA and not for DA and CA yet. In addition, some advanced gradient techniques are common for FA and DA methods, but not for CA methods. The Hessian matrix, for example, has been used to obtain second-order information in Integrated Hessians for FA and extensively in all IF-related methods for DA, and can be explored for CA.



Moreover, seeing the theoretical connections among FA, DA, and CA enables us to draw inspiration from one area to advance our understanding of another as a whole. For example, Section \textsection~\ref{sec:lfa} discusses how diverse FA methods all perform LFA. This framework may also apply to DA and CA. We know that FA performs function approximation of the blackbox model's predictions over the space of input features. One may hypothesize that DAs perform function approximation of the model's weights over the space of training data points and that CAs perform function approximation of the model's predictions over the space of model components. If so, function approximation may unify DA methods and unify CA methods as well.
As in the case of unifying FA methods, such theoretical unification may also provide many benefits to DA and CA, including conceptual coherence, elucidation of method properties, simplicity for theoretical analyses, and principled practical recommendations.

Another research direction is to move towards more holistic analyses of model behavior. These attribution methods provide insight into model behavior through different lenses: input features, training data, and model components. Each type of attribution provides different and complementary information about model behavior. For example, for a given model prediction, FA may not suggest that the model is relying on sensitive features to make predictions, but CA may uncover a set of neurons that encode biased patterns. In this sense, focusing only on one type of attribution, i.e., studying only one part of the model, is insufficient to understand model behavior. Thus, future research may develop approaches to enable more comprehensive model understanding, such as understanding how to use different types of attribution methods together, the settings under which different attribution types may support or contradict one another, and the interactions between the three model parts (e.g., how patterns in the training data are encoded in model neurons).

\subsection{Advancing research in broader AI}
\label{subsec:further_development}
Attribution methods also hold immense potential to benefit broader AI research for other applications. 
Especially with a unified view integrating FA, DA, and CA, researchers cannot only gain deeper insights of model behavior but also edit and steer models towards desired goals and improve model compliance with regulatory standards.

\textbf{Model editing}~\citep[inter alia]{de2021editing, mitchell2021fast, meng2022locating} focuses on precisely modifying models without retraining.
It enables researchers to correct model mistakes, analogous to fixing bugs in software.
This approach is particularly valuable for LLMs, which encode vast information in their parameters and are prohibitively expensive to retrain.
It can be viewed as a downstream task of attribution methods.
Once attribution methods locate an issue, editing methods can be applied to the problematic parts.
While editing aligns most closely with CA, other attribution types serve essential complementary functions.
FA identifies spurious correlations requiring correction, and DA helps select proper data points to induce efficient editing.
The unified attribution framework provides a holistic perspective that enables more efficient and accurate editing, especially when CA alone proves insufficient \citep{hase2024does}.

\textbf{Model steering}~\citep[inter alia]{zou2023representation,marks2023geometry,burger2024truth}
differs from model editing by integrating a steering vector into the model's inference process rather than modifying model parameters.
While editing focuses on specific knowledge modifications, steering guides model behavior at a higher level, such as enhancing truthfulness and harmlessness in LLMs.
Similar to model editing, a unified attribution framework can significantly enhance steering by better localizing target components to steer and generating more effective steering vectors through relevant features and training data.
Recently, steering methods that rely on more than one type of attribution methods have emerged~\cite{marks2023geometry,burger2024truth}.
In Appendix \ref{app:model_editing_and_steering}, we provide a more detailed discussion of how a unified attribution framework can enhance both model editing and steering and provide a illustrative example of steering for truthfulness of LLMs.

\textbf{Model regulation}~\citep[inter alia]{oesterling2024operationalizing}
is an emerging field examining the relationship between AI systems, policy, and societal outcomes. Regulation and policy frequently stress the need for transparency of AI systems as well as users' right to an explanation. 
Attribution methods provide an avenue for practitioners to ensure that AI systems meet these legal and ethical requirements, by providing information about the overall AI system as well as specific input-output behavior. 
FA reveals input processing patterns, DA exposes training data influences, and CA illuminates architectural roles.
This multi-faceted understanding enables more targeted and effective regulation.
For example, when addressing biased behavior, FA can be used to identify discriminatory input patterns, DA to trace problematic training samples or copyright infringements, and CA to locate architectural elements needing adjustment.
These complementary perspectives provide the comprehensive understanding needed to guide model regulation toward desired societal outcomes.

%% file: sections/alternative.tex
Several alternative views exist that (partially) oppose the unified perspective presented in this paper. We discuss their merits and connections to our work below.

One alternative view holds that different attributions represent distinct research areas that have benefited from independent evolution in the past, and although they share some ideas, they will continue to benefit from developing separately.
FA has historically been synonymous with "explainable AI," DA has been part of "data-centric AI" focusing on applications such as data selection, and CA has emerged as a branch of "mechanistic interpretability" aimed at reverse-engineering algorithms in foundation models.
This independent evolution has produced distinct motivations and best practices within each area~\citep{guidotti2018survey, hammoudeh2024training, bereska2024mechanistic}.
While this view does not strongly oppose unification, attribution methods can indeed be more easily compared with other works from their respective fields when viewed separately.
For example, FA can be compared with inherently interpretable models~\citep{koh2020concept} for feature importance, DA can be compared with data coresets~\citep{mirzasoleiman2020coresets} for data selection, and CA can be compared with model pruning~\citep{sun2023simple}.
Nevertheless, we demonstrate in the second part of our position that meaningful unification across attributions is possible and can inspire new research directions.
We emphasize the unified perspective and its benefits throughout this paper.

There are also alternative views on the taxonomy of attribution methods.
First, researchers have proposed other taxonomies within each type of attribution. For example, FA methods have been classified as either model-agnostic or model-specific~\citep{arrieta2020explainable}.
Although these categorizations aid method selection, they are not applicable across all attribution types.
CA methods, for instance, are inherently model-specific since they analyze defined model components.
Second, one might argue that our three-way categorization of FA, DA, and CA into perturbation-based, gradient-based, and linear-approximation methods does not apply to all attribution methods.
While this is true -- there are surrogate models for approximation beyond linear~\citep{sokol2020limetree} for FA (\Cref{app:fa_generalize}), simulation-based~\citep{chai2024training} and reinforcement learning-based~\citep{yoon2020data} methods for DA (\Cref{app:da_other}), and sparse autoencoder-based methods for CA~\citep{bricken2023towards} (\Cref{app:ca_generalize}) -- our categorization captures the majority of representative methods.
In summary, our unification extends beyond attribution-specific categorizations and highlights broader, cross-attribution similarities. It is relatively comprehensive and uncovers connections that remain obscured when examining attributions in isolation. For example, while methods from all three attribution types use gradients, they compute them for different purposes with different objectives.

Recent work by \citet{saphra2024mechanistic} offers seemingly another alternative view on attributions, highlighting that different communities assign distinct meanings to attributions, often guided by historical, cultural, and methodological differences. However, by recognizing that attributions carry community-specific definitions with epistemic commitments, rather than being strictly independent methodologies, this view is actually aligned with our unification position.

%% file: sections/conclusion.tex
In this paper, we present a unified view of attribution methods from explainable AI (e.g., FA), data-centric AI (e.g., DA), and mechanistic interpretability (e.g., CA). We demonstrate that, while these attributions evolved largely independently to explain different aspects of AI systems, they differ mainly in perspective (i.e., examining input features, training data, or model components) and share core techniques (i.e., perturbations, gradients, and linear approximations). This unified view not only bridges the fragmented landscape of attribution methods but also inspires new directions in interpretability research through addressing common challenges and cross-attribution innovation and in broader AI research through model editing, steering, and regulation.


%% file: sections/acknowledgment.tex
Usha Bhalla is supported by a Kempner Institute Graduate Research Fellowship.
We thank Suraj Srinivas, Yuanzhou Chen, Karim Saraipour, and Gaotang Li for their feedback on the draft of this work.

%% file: sections/appendix.tex
\section{Summary of notations}
\label{app:notations}
In Table~\ref{tab:notations}, we summarize the notation used in this paper.
\input{tables/notations}

\section{Summary of methods}
\label{app:common_methods}
As we discussed in the main paper, methods for feature, data, and component attribution can be categorized based on their technique: perturbation-based methods, gradient-based methods, and linear approximation methods.
Perturbation-based methods measure how a model's output changes when modifying specific elements, whether they are input features, training data points, or model components.
To capture interactions between multiple elements, all three types of attributions employ common mathematical tools, such as the Shapley value from game theory.
Gradient-based methods analyze model behavior by leveraging gradients to provide insights into the model's sensitivity to small input changes.
Gradients bridge model behavior and the elements we wish to attribute without perturbations.
Attributions are generated using different kinds of gradients: computing gradients of model outputs with respect to input features to quantify feature importance, calculating gradients of loss functions with respect to specific training data points to analyze data influence, or using gradients to approximate the effects of modifying model components.
Linear approximation methods fit linear models to approximate complex model behaviors.
The inputs to these linear models can be input features, training data points, or model components.
In some cases, binary indicators replace the actual elements as inputs to simplify the approximation.

In Table~\ref{tab:methods}, we summarize the attribution methods discussed in this paper, which we believe are the representative ones and align with the unified view we presented.
Most of the empty cells in Table~\ref{tab:methods} (labeled as ``--'') represent methods that may be promising but have not yet been explored in the literature, as discussed in \textsection~\ref{subsec:cross_aspect_inspiration}.
These represent research ideas that have been verified in one attribution type but remain unexplored in others.
The exception is mask learning for data attribution, which may be less promising because learning a high-dimensional mask of size \(n\) jointly with the model would be infeasible when the model has not been trained.

\section{Detailed discussion of feature attribution methods}
\label{app:fa}
Feature attribution methods can be broadly classified into three categories. 
Perturbation-based methods attribute feature importance by observing changes in model output when input features are altered or removed. These methods provide intuitive results but can be computationally expensive for high-dimensional data. 
Gradient-based methods utilize the model's gradients with respect to input features to attribute their importance. These methods are popular for differentiable models like neural networks, as they are often computationally efficient. 
Linear approximation methods construct interpretable linear models of input features that approximate the behavior of the original complex model in the vicinity of a specific input and compute attribution scores from the linear model coefficients. These methods offer a balance between interpretability and local accuracy. 
Each category of methods has its strengths and limitations, making the choice of method dependent on the specific model, data characteristics, and attribution requirements of the task at hand. We now extend the discussion of some methods mentioned in the main text in more detail and provide discussions of some additional methods.


\subsection{Direct perturbation for feature attribution}

\paragraph{RISE}~\citep{petsiuk2018rise}
is a direct perturbation method that addresses limitations of earlier methods like occlusion while expanding applicability to complex models.
The method provides a systematic approach for assessing feature importance through efficient sampling and aggregation of perturbations.
It operates by randomly masking different regions of the input image and measuring the model's output to each masked version.
The final saliency map is constructed by combining these random masks, with each mask weighted according to the model's predicted probability on the corresponding masked input.
This sampling-based approach allows RISE to efficiently estimate feature importance while capturing interactions between different image regions.

\subsection{Game-theoretic feature attribution}
\label{app:fa_game}
The Shapley value, a solution concept from cooperative game theory introduced by Lloyd Shapley~\citep{shapley1953value}, has gained particular prominence in feature attribution.
For a data point $\mathbf{x}$ with features $\{x_1, x_2, \ldots, x_d\}$, the Shapley value of feature $x_i$ for the model prediction $f(\mathbf{x})$ is defined as:

\[\phi_i(\mathbf{x}, f) = \sum_{x_S \subseteq \mathbf{x} \setminus \{x_i\}} \frac{1}{\binom{d-1}{|S|}} [f(x_S \cup \{x_i\}) - f(x_S)]\]

where $x_S$ represents a subset of features excluding feature $x_i$ indexed by $S$, and $f(x_S)$ denotes the model's prediction when only features in set $x_S$ are present.
$f(x_S \cup \{x_i\}) - f(x_S)$ is the marginal contribution of feature $x_i$ to the subset $x_S$ for the model's prediction.
The formula computes the average marginal contribution of feature $x_i$ across all possible feature subsets.
We simplify the attribution score notation by writing $\phi_i(\mathbf{x}) = \phi_i(\mathbf{x}, f)$.

Shapley values possess several desirable properties that make them particularly suitable for feature attribution:
\begin{itemize}
    \item Efficiency: The attributions sum to the total prediction, i.e., $\sum_{i} \phi_i(\mathbf{x}, f) = f(\mathbf{x}) - f(\emptyset)$
    \item Symmetry: Features that contribute equally receive equal attribution, i.e., if $f(x_S \cup \{x_i\}) - f(x_S) = f(x_T \cup \{x_j\}) - f(x_T)$ for all subsets $x_S, x_T \subseteq x$, then $\phi_i(\mathbf{x}, f) = \phi_j(\mathbf{x}, f)$
    \item Linearity: For models $f_1$ and $f_2$, $\phi_i(\mathbf{x}, a_1 f_1 + a_2 f_2) = a_1 \phi_i(\mathbf{x}, f_1) + a_2 \phi_i(\mathbf{x}, f_2)$ for constants $a_1$ and $a_2$.
    \item Null player: Features that don't affect the prediction receive zero attribution, i.e., if $f(x_S \cup \{x_i\}) - f(x_S) = 0$ for all subsets $x_S \subseteq x$, then $\phi_i(\mathbf{x}, f) = 0$   
\end{itemize}

These properties offer theoretical guarantees for fair and consistent feature attribution, making Shapley values a principled approach to understanding model behavior.
However, the exact computation requires evaluating $2^{|\mathbf{x}|}$ feature combinations, leading to various approximation methods in practice.


\paragraph{Other game-theoretic concepts}
Besides the Shapley value, other cooperative game-theoretic concepts are also applicable to feature attribution, offering different trade-offs between computational complexity and specific properties of the resulting attributions.
The Shapley Taylor Interaction Index (STII)~\citep{dhamdhere2019shapley} is another concept that can be used for feature attribution, which is a generalization of the Shapley value that explicitly considers interactions between features.
The Banzhaf Interaction Index (BII)~\citep{patel2021high} is particularly useful for considering joint feature interactions with simpler computation than the Shapley value.
The core value~\citep{yan2021if}, for instance, employs different axioms and emphasizes attribution stability.
Additionally, the Myerson value~\citep{chen2018shapley} and HN Value~\citep{zhang2022gstarx} are valuable when prior knowledge about the feature structure is available.
These alternative approaches provide researchers and practitioners with a range of tools to tailor their feature attribution methods to specific needs and constraints of their models and datasets.

\paragraph{Connection to linear approximation}
The most common Shapley value-based attribution method, SHAP~\citep{lundberg2017unified}, is a perturbation-based method rooted in cooperative game theory. 
However, it can also be viewed through the lens of linear approximation methods, representing a unified approach of local linear attribution and classic Shapley value estimation.
In the context of linear approximation, SHAP can be interpreted as fitting a linear model where features are players in a cooperative game, and the model output is the game's payoff.
The SHAP framework includes variants like Kernel SHAP, which uses a specially kernel for weighted local linear regression to estimate SHAP values, effectively approximating the model's behavior in the feature space surrounding the instance being explained.
This perspective on SHAP highlights its connection to linear approximation methods while retaining its game-theoretic foundations.

\subsection{Perturbation mask learning for feature attribution}
Mask learning methods offer several notable advantages in feature attribution.
They provide attributions more efficiently, especially for high-dimensional inputs, which is particularly beneficial for complex models and large datasets.
The continuous spectrum of importance scores generated by these techniques offers more nuanced insights than binary approaches, allowing for a finer-grained understanding of feature relevance.
Furthermore, the learning process can implicitly capture complex feature interactions, providing a more comprehensive view of how features contribute to model decisions.
Additionally, these methods can be tailored to specific model architectures and incorporate domain-specific constraints, enhancing their flexibility and applicability across various fields.


\paragraph{Masking model}~\citep{dabkowski2017real} pioneered the masking approach for image classification by introducing a masking model that generates pixel masks, aiming to identify a minimal set of features that sufficiently maintain the prediction of the original input. 
This approach employs a U-Net~\citep{ronneberger2015u} as the masking model, which first encodes the input image and then upsamples it to produce a mask.
The masking model acts as the attribution function \(g\), where mask values represent feature attribution scores.
While initial training is required, the masking model generates masks through a single forward pass at test time, which significantly improves runtime compared to earlier perturbation methods.
For learning the mask model, the main challenge is balancing feature minimality and predictive power.

\paragraph{L2X}~\citep{pmlr-v80-chen18j} frames feature attribution as an optimization problem and learns a masking model to generate masks that maximize mutual information between input feature subsets and model output.
This approach not only identifies important features but also captures their interdependencies, providing a comprehensive explanation of the model behavior.
L2X is versatile and applicable to various domains beyond image classification.
For instance, it has been successfully applied to sentiment classification tasks using datasets of movie reviews. 

\paragraph{Gradient computation in mask learning vs. gradient-based methods}
To avoid potential confusion, it is important to note that while mask learning methods may utilize gradient computation during the learning process, these gradients serve a different purpose than those in gradient-based attribution methods.
In mask learning, gradients are used to learn a soft mask or an explainer model for generating masks. These gradients are not used to directly determine the feature attribution scores themselves.
This distinction sets mask learning approaches apart from the gradient-based methods discussed in the following section.    

\subsection{Gradient-based feature attribution}
\label{app:gbfa}
Gradients for feature attribution are very different from those used in model training.
For a model \(f\) with parameters \(\theta\), gradients of the loss function are taken with respect to the parameters (\(\nabla_\theta \mathcal{L}(\theta)\)) to guide parameter updates during training.
In feature attribution, gradients of the model's output \(f(\mathbf{x})\) are taken with respect to input features (\(\nabla_{\mathbf{x}} f(\mathbf{x})\)) to quantify each feature's contribution to the model's output.
Gradient-based methods have been widely adopted for feature attribution because of their computational efficiency.
They typically require only a single forward and backward pass through the model to compute the gradients, and require no additional perturbation or linear model fitting, which makes them particularly suitable for real-time applications and large-scale datasets.
On the other hand, gradient-based methods have two key limitations.
They require access to model parameters and only work with differentiable models.
Additionally, the gradient results can be nonrobust as we discussed in Section~\ref{subsec:common_challenges}.
Despite these challenges, gradient-based methods remain a fundamental tool in the feature attribution toolkit.

\paragraph{Gradient $\times$ Input}~\citep{shrikumar2017learning} improves over vanilla gradients. By multiplying the input features element-wise with their corresponding gradients, this method mitigates the ``gradient saturation'' problem where gradients can become very small even for important features. The element-wise multiplication also helps reduce visual diffusion in the attribution score visualization, resulting in sharper and more focused visualizations of important features. 

\paragraph{Integrated Gradients}~\citep{sundararajan2017axiomatic} provides a theoretically grounded approach to feature attribution by accumulating gradients along a path from a baseline input to the actual input. This method satisfies important axioms including sensitivity (a change in input leads to a change in attribution) and implementation invariance (attributions are identical for functionally equivalent networks). The integration process captures the cumulative effect of each feature as it transitions from the baseline to its actual value, providing a more complete picture of feature importance than vanilla gradients of a single input.

\paragraph{Integrated Hessians}~\citep{janizek2021explaining} extends the integrated gradients method to analyze feature interactions, with the goal of understanding how features interact.
This method treats the integrated gradient function as differentiable and quantifies interactions between two features using second-order information, the Hessian matrix.
By computing these Hessian-based interactions along the same integration path used in integrated gradients, it provides a principled way to measure feature interdependencies and more comprehensive feature attributions.

\paragraph{Guided Backpropagation (GBP)}~\citep{springenberg2014striving} modifies the standard backpropagation process of NNs to generate cleaner and more interpretable attribution results. When propagating gradients through ReLU units, GBP sets negative gradient entries to zero, effectively combining the signal from both the higher layer and the ReLU units. This modification helps eliminate artifacts and noise in the attribution results while preserving the positive contributions of features, resulting in sharper and more visually interpretable feature attributions.

\paragraph{Grad-CAM}~\citep{selvaraju2016grad} is a widely used method for attribution and visualization of important regions in images, specifically for convolutional NNs. It computes the gradient of the target class score (logit) with respect to the feature maps in the last convolutional layer. These gradients are then used as weights to combine the feature maps, creating a coarse localization map that highlights important regions for predicting the target class. The resulting localization map is upsampled to match the input image size to create a saliency map, providing an interpretable visualization of features important for the model's output.

\paragraph{Generalizing gradient-based feature attribution} Several additional methods share similar underlying principles with gradient-based approaches, although they do not directly compute gradients in their original formulation.
For example, Layer-wise Relevance Propagation (LRP)~\citep{bach2015pixel} propagates predictions backwards through the network while preserving the total relevance at each layer.
LRP provides a unique perspective on attribution by focusing on the relevance of individual neurons to the final prediction.
Similarly, DeepLIFT~\citep{shrikumar2017learning} operates by comparing each neuron's activation to a reference activation and propagating the resulting differences to the input features.
Interestingly, \citet{ancona2017unified} demonstrated that for ReLU networks with zero baseline and no biases, both $\epsilon$-LRP and DeepLIFT (rescale) methods are mathematically equivalent to the Input $\times$ Gradient approach.
For a more detailed analysis of these equivalences, we refer readers to \citet{ancona2017unified}.

\subsection{Linear approximation for feature attribution}

\paragraph{C-LIME}~\citep{agarwal2021towards} is a variant of LIME specifically designed for continuous features that generates local explanations by sampling inputs in the neighborhood of a given point.
It differs from LIME in several aspects: it uses a constant distance metric and Gaussian sampling centered at the input point rather than uniform random sampling, making perturbations naturally closer to the input without requiring explicit weighting.
C-LIME also restricts itself to linear models for continuous features, unlike LIME's more general model class, and excludes regularization by setting the regularizer term to zero.
For simplicity, C-LIME focuses on feature weights while ignoring the intercept terms.

\paragraph{Generalizing linear approximations for feature attribution}
\label{app:fa_generalize}
For both LIME, C-LIME, and other similar linear approximation methods, the assumption that the model's behavior can be reasonably approximated by a linear function in the local neighborhood is crucial.
In this context, while the linear model serves as a proxy, it can be replaced by a more complex yet interpretable model that is still capable of providing attribution results. For instance, LIMETree uses tree models~\citep{sokol2020limetree}. We refer readers to~\citet{sokol2020limetree} for a detailed discussion of this approach.

\subsection{Unifying feature attribution methods through local function approximation}
\label{app:unify-fa}

\newcommand{\E}{\mathop{\mathbb{E}}}

Under the local function approximation framework, the model $f$ is approximated by an interpretable model class $\mathcal{G}$ around the point of interest $\mathbf{x}$ over a local neighborhood distribution $\mathcal{Z}$ using a loss function $\ell$. The approximation is given by 
$$g^{*} = \mathop{\arg\min}\limits_{g \in \mathcal{G}} \mathop{\mathbb{E}}\limits_{\xi \sim \mathcal{Z}} \ell(f, g, \mathbf{x}, \xi).$$
\citet{lfa2022} show that at least eight feature attribution methods (Occlusion, KernelSHAP, Vanilla Gradients, Gradients $\times$ Input, Integrated Gradients, SmoothGrad, LIME, and C-LIME) are all instances of this framework. 
These methods all use the linear model class $\mathcal{G}$ to approximate $f$, but do so over different local neighborhoods $\mathcal{Z}$ using different loss functions $\ell$ as in Table~\ref{tab:lfa}. 

Under this setup, $g$'s model weights are equivalent to the explanation obtained using each method's original algorithm. 
Also, note that for the local function approximation framework, there are requirements on the loss function: a valid loss $\ell$ is one such that $\E_{\xi \sim \mathcal{Z}} \ell(f,g,\mathbf{x},\xi) = 0 \iff f(\mathbf{x}^{\{\xi\}}) = g(\mathbf{x}^{\{\xi\}})~~~\forall \xi \sim \mathcal{Z}$.

In addition, while the local function approximation framework may seem similar to LIME, it differs from LIME by 1) requiring that $f$ and $g$ share in the same input and output domain, 2) imposing the condition on the loss function $\ell$ discussed above, and 3) following the standard machine learning methodology to avoid overfitting and to tune hyperparameters. A more detailed discussion can be found in Section 3 of \citet{lfa2022}.

\input{tables/lfa.tex}

\section{Detailed discussion of data attribution methods}
\label{app:da}
Unlike feature attribution methods, data attribution methods examine the training phase and quantify the training data's influence on the model's output. However, data attribution methods can be categorized into the same three categories.
Perturbation-based methods assess training data importance by observing changes in model behavior when training samples are removed or modified. 
These methods provide accurate results but can be computationally expensive as they require retraining the model multiple times. 
Complete retraining-based methods like LOO are often used as a ground truth for evaluating other data attribution methods.
Gradient-based methods utilize the model's gradients evaluated at the training data points and the test data point to quantify the influence of the training data points on the test data point. 
These methods avoid the computational cost of retraining the model but may face the challenges like the non-convexity of the loss landscape or the difficulty in computing the Hessian matrix efficiently. 
Linear approximation methods construct interpretable models that approximate how training data affects model behavior. 
The linear model operates on the entire training dataset, which can be suprisingly accurate but also heavy to train.
Methods from different categories have their own strengths, limitations, and use cases.
We now extend the discussion of some prominent methods mentioned in the main text and discuss some additional methods.

\subsection{Leave-One-Out data attribution}

Leave-One-Out is a prominent example of perturbation-based data attribution, but also a natural idea that existed for a long time in statistics.
For example, it has been used as a resampling technique (e.g., jackknife resampling ~\citep{tukey1958bias}) to estimate the bias and variance of a statistic of interest (such as a regression coefficient). 
LOO has been used to detect influential data points for linear regression~\citep{cook1982residuals}, for example, through Cook's distance~\citep{cook1977detection}.
Until recently, LOO has been applied to modern AI models to attribute model performance to individual training data points~\citep{jia2021scalability}.
It provides valuable counterfactual insights with its main limitation being computational cost, as it requires retraining the model for each data point.
Many newer attribution methods, like the gradient-based methods can be viewed as efficient approximations of LOO.

\subsection{Game-theoretic data attribution}
\label{app:da_game}
The primary limitation of game-theoretic methods is their prohibitive computational cost for large datasets, as they require numerous model retrainings over the powerset of the training data.
To address this challenge and further improve method robustness, researchers have proposed various methods.

\paragraph{Truncated Monte Carlo (TMC) Shapley}~\citep{ghorbani2019data} approximates the Shapley value by adopting an equivalent definition of the Shapley value in terms of aggregating over data permutations instead of data subsets~\citep{shapley1953value}. The method works by truncating the number of permutations sampled and the number of data points considered in each permutation, and align that with model training. For each sampled permutation, it computes the marginal contribution of each data point by evaluating model performance with and without that point. The gradient information from these evaluations is then used to as an estimate of each point's marginal contribution. TMC Shapley significantly reduces computational cost while maintaining reasonable approximation accuracy of the exact Shapley values.

\paragraph{KNN Shapley}~\citep{jia2019efficient} introduces an efficient approximation for Shapley values data attribution by using K-Nearest Neighbors (KNN) as a surrogate model instead of retraining the full model. For each test point, it first identifies its K nearest neighbors in the training set. Then, it computes Shapley values only considering these neighbors' contributions to the KNN prediction, rather than the original model's prediction. This localized computation dramatically reduces complexity from exponential to polynomial in the number of neighbors K. The method maintains good attribution quality since nearby training points typically have the biggest influence on a test point. The KNN approximation aligns better with the goal of estimating the value of data from the data vendor's perspective, and thus was named data valuation in the original paper.

\paragraph{Beta Shapley}~\citep{kwon2022beta} extends the standard Data Shapley framework by introducing a beta distribution to weight different subset sizes differently. The original Shapley value weights subsets according to their sizes. The $\beta$ parameter controls how much emphasis is placed on smaller versus larger subsets when computing marginal contributions. This generalization relaxes the efficiency axiom of classical Shapley values, which requires attributions to sum to the total model value. By allowing this flexibility, Beta Shapley can better handle noisy or corrupted training data by reducing their influence on the attribution scores. The method provides theoretical analysis showing how different $\beta$ values affect properties like noise robustness and estimation variance.

\paragraph{Data Banzhaf}~\citep{wang2023data} adapts the Banzhaf value from cooperative game theory as an alternative to Shapley values for data attribution. The Banzhaf value considers the average marginal contribution of training point across all possible data subsets like the Shapley value, but weights these contributions differently. This weighting scheme leads to the largest possible safety margin, making the attribution more robust to data perturbations and noise. The method provides theoretical guarantees on this robustness and demonstrates empirically that it can better identify mislabeled or adversarial training examples compared to Data Shapley.

\subsection{Influence function and its variants}
\label{app:da_if}

Influence functions provide a way to estimate how model parameters would change if we reweight or remove a training point, without having to retrain the model.
Given a model with parameters \(\theta\) trained by minimizing the empirical risk \(\frac{1}{n}\sum_{i=1}^n \mathcal{L}(\mathbf{x}^{(i)}, \theta)\) over the training, the IF approximates the change in parameters when upweighting a training point \(\mathbf{x}^{(j)}\) by \(\epsilon\):
\[\theta_{\epsilon, \mathbf{x}^{(j)}} = \argmin_{\theta} \frac{1}{n}\sum_{i=1}^n \mathcal{L}(\mathbf{x}^{(i)}, \theta) + \epsilon \mathcal{L}(\mathbf{x}^{(j)}, \theta)\]
Under the assumption that the loss function \(\mathcal{L}\) is twice-differentiable and strictly convex, a first-order Taylor expansion around the final optimal model parameters \(\theta^*\) gives:
\[\mathcal{I}_{\text{up,params}}(\mathbf{x}^{(j)}) = -H_{\theta^*}^{-1} \nabla_\theta \mathcal{L}(\mathbf{x}^{(j)}, \theta^*)\]
where \(H_{\theta^*} = \frac{1}{n}\sum_{i=1}^n \nabla_\theta^2 \mathcal{L}(\mathbf{x}^{(i)}, \theta^*)\) is the Hessian and is by assumption positive definite.
The influence of training point \(\mathbf{x}^{(j)}\) on the loss at test point \(\mathbf{x}^{\text{test}}\) is the effect of this infinitesimal \(\epsilon\)-upweighting on test point’s risk:
\[\mathcal{I}_{\text{up,loss}}(\mathbf{x}^{(j)}, \mathbf{x}^{\text{test}}) = - \nabla_\theta \mathcal{L}(\mathbf{x}^{\text{test}}, \theta^*)^\top H_{\theta^*}^{-1} \nabla_\theta \mathcal{L}(\mathbf{x}^{(j)}, \theta^*)\]
The negative \(\mathcal{I}_{\text{up,loss}}(\mathbf{x}^{(j)}, \mathbf{x}^{\text{test}})\) will be the data attribution score \(\psi_j(\mathbf{x}^{\text{test}})\) on \(\mathbf{x}^{\text{test}}\) and it provides an efficient approximation to LOO retraining~\citep{koh2017understanding}.
 
While effective in certain scenarios and computationally more feasible than retraining-based methods, IF faces several challenges. 
First, it assumes convexity and double-differentiability, which are often not satisfied in deep learning scenarios.
Second, it involves Hessian matrix computation, which can be computationally expensive for large models.
Also, the potential non-positive definiteness of the Hessian matrix in certain cases can lead to inaccuracies, often necessitating the introduction of dampening factors that may affect the precision of influence estimates.
To address these limitations, many methods have been proposed to enhance the efficiency and applicability of IF.

\paragraph{FastIF}~\citep{guo-etal-2021-fastif} introduces several key optimizations to make IF more computationally tractable. First, it uses KNN to reduce the search space from the entire training set to a smaller subset of promising candidates that are likely to be influential. Second, it develops a fast estimation technique for the inverse Hessian-vector product that avoids computing and storing the full Hessian matrix and its inverse. Third, it implements parallelization strategies to asynchronously compute Hessian-vector products across multiple processors. These optimizations together enable FastIF to scale to much larger datasets while maintaining attribution quality comparable to the original IF.

\paragraph{Arnoldi IF}~\citep{schioppa2022scaling} employs Arnoldi's iterative algorithm to efficiently identify the dominant eigenvalues and eigenvectors of the Hessian matrix. These dominant components serve as the basis for projecting all gradient vectors into a lower-dimensional subspace. Compute IF in this subspace substantially reduces the computational complexity. The method can be flexible by selecting an appropriate number of eigenvalues to retain. Empirical results demonstrate that this approach can achieve comparable attribution quality to full IF while significantly reducing both memory requirements and computation time.

\paragraph{EK-FAC}~\citep{grosse2023studying} leverages the Eigenvalue-corrected Kronecker-Factored Approximate Curvature (EK-FAC) parameterization to efficiently approximate the Hessian matrix. This parameterization exploits the natural block structure present in NNs to decompose the Hessian into more manageable components, which significantly reduces the computational complexity of Hessian-vector products. By leveraging these techniques, IF can be effectively scaled to large transformer models with hundreds of millions of parameters, which are orders of magnitude more complex than the simpler NNs originally considered by~\citet{koh2017understanding}. Theoretical guarantees for the approximation quality and demonstrations of empirical success on foundation models were shown.

\paragraph{RelateIF}~\citep{barshan2020relatif} addresses another limitation of IF other than their computational cost. Standard IF methods often highlight outliers or mislabeled data points as most influential, which may not always align with intuitive notions of influence. RelateIF introduces a novel approach that distinguishes between global and local influence by examining how training data affect specific predictions relative to their overall impact on the model. This relative influence measure helps identify training data points that have significant local influence on particular test predictions while accounting for their broader effects on the model. RelateIF better captures intuitive notions of influence while being more robust to outliers in the training data.

\subsection{Tracing training path for data attribution}
\label{app:da_dynamic}

Tracing training path for data attribution provides valuable insights of training dynamics while avoiding limitations of LOO and IF. These methods that trace training dynamics that provide more accurate attribution results but they are all more computationally expensive than those considering only final model parameters like IF.

\paragraph{TracIn}~\citep{pruthi2020estimating} traces the data influence throughout the entire training process.
The method attributes influence by computing dot products between training and test data gradients at each training step from the initial model parameters to the final model parameters at the end of training, accumulating these to capture a training point's total influence across the training path.
This path tracing approach provides valuable insights into training dynamics while avoiding limitations of LOO and IF, such as assigning identical attribution scores to duplicate training data points.
TracIn also offers greater flexibility than IF by eliminating the convexity assumption and Hessian matrix computations.
On the other hand, its tracing requires storing intermediate model checkpoints during training, increasing both memory usage and computational costs.

\paragraph{SGD-Influence}~\citep{hara2019data} traces the training path by approximating the training process with a series of unrolled steps to estimate data influence.
The method estimates LOO influence by unrolling gradient descent using empirical risk Hessians, under the assumption that both the model and loss function are convex and the optimization algorithm is Stochastic Gradient Descent (SGD).
SGD-Influence primarily applies unrolling to quantify the Cook's distance~\citep{cook1977detection} between model parameters with and without a specific training point.
To better align with attribution estimation, a surrogate linear influence estimator is used to incrementally update throughout the unrolling process.
However, this approach requires unrolling the full training path for each test instance individually, which has significant computational complexity.

\paragraph{SOURCE}~\citep{bae2024training} extends training path tracing to better capture the training dynamics and reduce the computational cost.
It bridges the gap between gradient-based approaches like IF and unrolling-based methods like SGD-Influence.
While IF is computationally efficient, it struggles with underspecification of the training dynamics.
Unrolling-based methods address these limitations but face scalability challenges.
SOURCE combines the benefits of both approaches by using an IF-like formula to compute approximate unrolling.
This makes SOURCE both computationally efficient and suitable for scenarios where IF struggles, such as non-converged models and multi-stage pipelines.
Empirically, SOURCE demonstrates superior performance in counterfactual prediction compared to existing data attribution methods.

\subsection{Linear approximation and Datamodel}
\label{app:da_linear}

\paragraph{Datamodel}~\citep{ilyas2022datamodels} applies linear approximation to DA, similar to LIME in FA.
It constructs a linear model \(g\) with \(n\) coefficients and \(\{0,1\}^n\) vectors as inputs, where each input represents a subset of training data.
\(g\) is learned to map any counterfactual subset of training data to output \(f(\mathbf{x})\), where \(f\) is trained on this subset with the given model architecture and training algorithm.
The coefficients of \(g\) thus represent the attribution scores of the training data points.
The method's counterfactual nature enables evaluation of other attribution methods via the \textit{Linear Datamodeling Score (LDS)}, which compares their attribution score rankings to Datamodel's ranking.
While Datamodel can effectively capture model behavior, constructing this large linear model requires extensive counterfactual data obtained by training model \(f\) on various subsets, making it computationally intensive.

\paragraph{TRAK}~\citep{park2023trak} addresses the computational challenges of Datamodel by estimating Datamodels in a transformed space where the learning problem becomes convex and can be approximated efficiently.
It further improves efficiency through random projection of model parameters and ensemble attribution results of multiple trained models.
Though the ensemble approach still requires some model retraining on different subsets, it achieves high estimation accuracy with significantly fewer retraining iterations than Datamodel.
Furthermore, both approaches can be viewed as perturbation-based methods, similar to LIME, as they 
systematically vary training data to construct linear models.

\subsection{Data attribution methods using other techniques}
\label{app:da_other}

There are data attribution methods that use other techniques beyond the three core techniques we discussed in the main paper to estimate the influence of each training data point on the model's output.

\paragraph{GPTfluence}~\citep{chai2024training} is a simulation-based method that estimates how individual training examples influence the performance of GPT models. GPTfluence uses a featurized simulation-based approach to model the training trajectory. It begins by collecting training dynamics across multiple runs, including sequences of training batches and performance metrics like loss, BLEU, and ROUGE scores. GPTfluence then trains a simulator that represents both training and test examples using frozen pre-trained embeddings (e.g., from BERT), and learns multiplicative and additive influence factors through inner products between these embeddings. The simulator models performance at each step using an n-th order Markov process that conditions on previous steps, allowing it to predict the full evolution of test metrics over training. 

\paragraph{Data Valuation using Reinforcement Learning (DVRL)}~\citep{yoon2020data} uses reinforcement learning to estimate the influence of each training data point on the model's output. DVRL jointly trains a data value estimator and a task predictor model. At each iteration, the data value estimator assigns selection probabilities to training examples, determining which examples are used to update the predictor model. The quality of these selections is evaluated using a small validation set, and a RL signal based on the validation performance is used to update the data value estimator. This allows DVRL to dynamically learn which samples are most useful for improving the model’s performance.

\section{Detailed discussion of component attribution methods}
\label{app:ca}

Unlike feature and data attribution methods, component attribution methods analyze the internal mechanisms of models by attributing model behavior to specific architectural components like neurons, layers, or attention heads. These methods similarly fall into three categories.
Perturbation-based methods assess component importance by observing changes in model behavior when specific components are modified, resulting in various forms of causal mediation analysis.
Gradient-based methods utilize gradients with respect to component activations to approximate the component importance in causal mediation analysis.
Linear approximation methods construct linear models that directly approximate how components affect model behavior.
Methods from different categories have their own strengths and limitations.
We now extend the discussion of some prominent methods mentioned in the main paper and provide discussions of some additional methods.

\subsection{Perturbation-based component attribution} 

\paragraph{Various types of ablations in causal mediation analysis} 
Causal mediation analysis is frequently referred to as activation patching, wherein activations of the specific component from the clean run are patched into the corrupted run to ascertain if those activations are sufficient and necessary to retrieve the desired output. 
Activation perturbations can consist of zero ablations \citep{olsson2022context, geva2023dissecting}, mean ablations \citep{wang2022interpretability}, smoothed Gaussian noising \cite{meng2022locating}, interchange interventions \citep{geiger2021causal}, learned ablations \citep{li2024optimal}.
In all cases, the dataset used to generate the activations must be chosen to elicit the desired model behavior, with a matching metric that measures the success of the behavior.

\paragraph{Automated Circuit Discovery (ACDC)}
Similar to subnetwork pruning, ACDC \citep{conmy2023towards} is tries to find a subnetwork that is far sparser than the original graph and recovers good performance on the task. This is done by iterating through the computational graph of the model from outputs to inputs and attempting to remove as many edges between nodes as possible without reducing the model's performance. In this case, performance is measured as the KL-divergence between the full model and the subgraph’s predictions. Furthermore, masked or ablated edges are replaced with activations from a corrupted run or counterfactual input prompt, rather than zero-ablated as is done in subnetwork probing. 

\subsection{Gradient-based component attribution} 

\paragraph{Edge Attribution Patching (EAP)}
\citep{syed2023attribution} combines ACDC and Attribution Patching to create EAP, which generates attribution scores for the importance of all edges in the computational graph through normal attribution patching and then sorts those scores to keep only the top $k$ edges in a circuit, thus yielding the circuit corresponding to the task.

\subsection{Generalizing the definitions of components}
\label{app:ca_generalize}

While initial works explored this form of causal mediation analysis where each neuron was an individual component \citep{bau2020understanding, vig2020investigating}, recent work has moved towards other mediators due to the computational intractability of considering individual neurons in larger models and due to hypotheses of entanglement and polysemanticity of neurons in foundation models. Furthermore, recent work has argued that specific mediators are only reasonable for certain behaviors \citep{mueller2024quest} and have also explored the feasibility of patching activations both within and between models to increase expressivity \citep{ghandeharioun2024patchscope}. 

\paragraph{Sparse Autoencoders (SAEs)}~\citep{bricken2023towards, cunningham2023sparse}
are trained to reconstruct model activations under sparsity constraints.
Through learning sparse, overcomplete representations of model activations, SAEs effectively decompose complex, entangled features into more interpretable components.
The enforced sparsity ensures that each SAE feature captures a distinct and meaningful aspect of the model's behavior, making them useful for model understanding.
Recent research has demonstrated that SAEs can successfully extract interpretable components, but since SAEs focus on learning new components rather than attributing to existing ones in the original model, we do not consider them as strictly component attribution methods in this paper. They can rather serve as a technique for discovering interpretable features that can subsequently be used for attribution. As the next paragraph shows, SAEs can be used for component attribution by first discovering interpretable components and then using them for attribution.

\paragraph{Sparse Feature Circuits}\citep{marks2024sparse}
Sparse feature circuits build upon the gradient-based attribution method attribution patching to determine the linear directions relevant to the task or behavior of interest. This method leverages sparse autoencoders to find directions in the models’s latent space that correspond to human-interpretable features. They then employ linear approximations similar to attribution patching, using either input gradients or integrated gradients, to efficiently identify which of the learned sparse autoencoder features are most relevant to the model behaviors, as well as connections between these features.

\section{Overcoming attribution challenges through a unified lens}
\label{app:common_challenges}

\subsection{Improving computational efficiency}
\label{app:dis_computation}

Computation challenges present substantial barriers that often prevent attribution methods from being applied to large-scale AI models, such as the foundation models with billions of parameters.
For perturbation-based methods, the curse of dimensionality makes a comprehensive analysis intractable when the number of required perturbations is large. For example, the full power set perturbation. 
This holds for all three types of attributions including high-dimensional inputs, large training datasets, and models with numerous components.
Game-theoretic methods face particular difficulties, as exact computation of the Shapley value is often prohibitively expensive and requires approximation techniques like Monte Carlo sampling.
The computational burden is also severer for data attribution methods, which require model retraining for each perturbation.
Gradient-based methods are more practical for large-scale models.
However, gradients essentially only provide first-order approximations of model behavior, which are inadequate to capture complex model behaviors and more sophisticated gradient formulations are needed for better attribution results.
For instance, TracIn require aggregating gradients across multiple stages, while IF demand computation of second-order Hessian matrices, leading to increased computational overhead.
Linear approximation methods also face computational hurdles in achieving high-quality approximations.
Model behavior can be complex, requiring numerous data points and model evaluations to establish sufficient data for learning accurate linear models.
Furthermore, for all three types, most attribution methods must compute results separately for each new test data point, creating additional computational strain when attribution analysis is needed for large datasets.
Therefore, improving computational efficiency calls for better sampling methods to avoid perturbing the entire input space in perturbation-based methods and to learn more accurate linear models for linear approximation methods. Also needed are more efficient gradient computations, especially for higher-order gradients like Hessians. Many works have sought to improve these aspects, with promising examples like \citet{grosse2023studying} showing that gradient-based methods can be scaled to large language models (LLMs) with up to 52 billion parameters. However, there is still much room for improvement to make attribution methods more efficient.

\subsection{Improving attribution consistency}
\label{app:dis_consistency}

The consistency problem in attribution methods is a significant concern. 
This challenge is also prevalent across due to variability introduced in sampling, learning processes with stochastic optimization, and also non-trivial hyperparameters.
When attribution involves sampling, such as the Monte Carlo sampling in some perturbation-based methods to avoid the full power set perturbation, the inherent randomness leads to varying attribution results.
Besides, when attribution involves learning processes with stochastic optimization, as seen in mask-learning perturbation and linear approximation methods, different learning outcomes yield inconsistent attribution results.
Many attribution methods rely on hyperparameters that can lead to different attribution outcomes.
These include sampling parameters, such as the number of samples used for computing Shapley values.
They also include optimization hyperparameters for various learning approaches, such as learning rates and number of steps in linear approximation methods.
Additionally, approximation hyperparameters are needed for quantities that are computationally challenging to calculate directly, such as dampening factors for inverse Hessian-vector products.
Further variability is introduced through fundamental design choices, such as the selection of perturbation type in perturbation-based methods, where options include mean perturbation, zero perturbation, and random perturbation.
While these different approaches should theoretically produce similar results based on their underlying principles, in practice they often yield notably different attributions.
While some gradient-based methods can produce consistent results in a single run when they do not involve sampling or approximations for computationally intensive operations, the consistency among different gradient-based methods varies considerably, which is also a problem for all three techniques and leads to the following evaluation challenges.
We see the potential improvement for this challenge including producing more deterministic attribution methods that are less sensitive to random hyperparameters, provide provable guarantees of attribution stability or error bounds. While past works have made some progress in this direction, there were strong assumptions added to the setting, e.g., convexity~\citep{koh2017understanding}. Thus, more efforts are needed to improve consistency in more realistic settings.


\subsection{Towards fairer and more practical evaluation}
\label{app:dis_eval}

Evaluating attribution methods presents significant challenges due to the lack of ground truth and the inherent complexity of modern AI systems.
These challenges stem partially from the inconsistency problem, the computational cost and generalizability of some evaluation metrics, and the lack of universal definitions of importance and ground truth.
These common evaluation approaches and their limitations are summarized below.

\paragraph{Counterfactual evaluation} is a widely used approach that assesses attribution methods by comparing their scores with the actual impact of removing or modifying elements.
Common metrics include \textit{fidelity}, which evaluates sufficiency by retaining only elements with high attribution scores while removing those with low scores.
Conversely, \textit{inverse fidelity} measures necessity by removing elements with high attribution scores while retaining those with low scores.
LOO attribution represents a special case of inverse fidelity.
For data attribution specifically, more sophisticated metrics like LDS for data attribution compare attribution rankings with the actual impact of removing training data points, with LDS being a sophisticated case of fidelity.
An important implicit metric in counterfactual evaluation is \textit{sparsity} or \textit{minimality}, which measures how few elements are needed to achieve high fidelity.
Greater sparsity is desirable as it indicates that fewer elements are required for explanation.
While counterfactual evaluation provides concrete validation, it faces two major challenges:
The computational cost of generating counterfactuals, particularly for data attribution, can be prohibitive.
Additionally, the complex interactions between elements may not be fully captured by individual counterfactual evaluations.

\paragraph{Task-specific evaluation} assesses the practical utility of attribution methods in downstream tasks.
For instance, feature attribution can help identify feature changes that can flip model outputs, while data attribution scores can detect mislabeled training examples, and component attribution scores can help identify the most important components that allows for model pruning.
Attribution methods can be compared based on the performance on these specific tasks.
While this approach provides practical validation, its findings may not generalize effectively across different tasks or domains.

\paragraph{Human evaluation} relies on domain experts or users to assess the quality and interpretability of attributions.
This approach is especially valuable for validating whether attributions align with human understanding and domain expertise.
For example, for feature attributions, the attribution results can be considered if they generate clearer visual saliency maps that align with human intuition.
While human evaluation provides valuable real-world validation, it can be both subjective and resource-intensive and can only be treated as the gold standard in certain cases.

The development of more robust and comprehensive evaluation frameworks remains a crucial research direction for advancing all attribution methods.

\section{Advancing attribution benefits broader AI research}

\paragraph{Model editing and steering}
\label{app:model_editing_and_steering}
A unifying framework offers a principled foundation and enables coordinated use of different attribution types—for example, combining CA’s localization with DA’s data selection under consistent semantics—leading to more effective and generalizable interventions. While recent model editing and model steering approaches have started to selectively target features, data, or components for better results, these methods are often heuristic. A unified attribution perspective can make these decisions more systematic and effective.

We use model steering for truthfulness as an illustrative case. In the foundational RepE work~\cite{zou2023representation}, steering is performed via Low-Rank Representation Adaptation (LoRRA) trained on a set of contrastive input pairs. In RepE, 1) The contrastive dataset is constructed using random samples from QA datasets, 2) the layer to steer is treated as a hyperparameter, and 3) the target token position to steer is implicitly the final token. We argue that DA, CA, and FA can each contribute to a more principled approach to these three steps respectively, through selecting contrastive examples, localizing the appropriate layer, and identifying the right token position. While the original choices in RepE are not necessarily flawed (e.g., steering at the last token might give the best results), a unified attribution perspective offers a more principled framework with room for improvement.

Subsequent work supports this idea. For instance, \citet{marks2023geometry} applies causal tracing to identify the most intuitive layer and token position for steering—an application of CA/FA ideas. Furthermore, \citet{burger2024truth} shows that the steering direction identified in \citet{marks2023geometry} is not universal: models trained on affirmative vs. negated statements yield different truthfulness directions. Recognizing this limitation essentially involves a manual form of DA.

While the improvements in \citet{marks2023geometry} and \citet{burger2024truth} are already promising, we believe that a unified attribution view can further enhance them. For example, the main-mass method used in \citet{marks2023geometry} and \citet{burger2024truth} identifies steering directions by cluster centers, requiring application to a whole dataset. In contrast, principled data selection via DA could find effective steering directions with fewer samples. Importantly, these selected samples will be most effective when they reflect the truthfulness through critical components and features—highlighting the need for a holistic attribution approach.

%% file: tables/notations.tex
\begin{table}[htb]
    \caption{Summary of notations.}
    \small
    \centering
    \begin{tabular}{ll}
    \toprule
    \textbf{Notation} & \textbf{Description} \\
    \midrule
    \(\mathcal{D}_{\text{train}}\) & Training dataset \(\{\mathbf{x}^{(1)}, \cdots, \mathbf{x}^{(n)}\}\) \\
    \(f_\theta\)/ \(f\) & Model trained on \(\mathcal{D}_{\text{train}}\), parameters \(\theta\) may be omitted \\
    \(c\) & Internal model components \(\{c_1, \cdots, c_m\}\), definition is method-specific \\
    \(\mathbf{x}^{\text{test}}\)/\(\mathbf{x}\) & Model input at test time for inference, superscript ``test" may be omitted \\
    \(x_i \) & The \(i\)-th input feature out of \(d\) features \\
    \(\phi_i(\mathbf{x})\) & Attribution score of input feature \(x_i\) for model output \(f(\mathbf{x})\) \\
    \(\psi_j(\mathbf{x})\) & Attribution score of training data point \(\mathbf{x}^{(j)}\) for model output \(f(\mathbf{x})\) \\
    \(\gamma_k(\mathbf{x})\) & Attribution score of internal model component \(c_k\) for model output \(f(\mathbf{x})\) \\
    \(g\) & Attribution function, which provides attribution scores for elements \\
    \(\mathcal{L}\) & Loss function for training the model \(f\) \\
    \(\ell\) & Loss function for learning the attribution function \(g\) \\
    \bottomrule
    \end{tabular}
    \label{tab:notations}
\end{table}

%% file: tables/lfa.tex
\begin{table}[t]
    \centering
    \caption{Existing methods perform local function approximation of a black-box model $f$ using the interpretable model class $\mathcal{G}$ of linear models where $g(\mathbf{x}) = \mathbf{w}^\top \mathbf{x}$ over a local neighbourhood $\mathcal{Z}$ around point $\mathbf{x}$ based on a loss function~$\ell$. $\odot$ indicates element-wise multiplication. (Table reproduced from \citet{lfa2022}).}
    \resizebox{\textwidth}{!}{
    \begin{tabular}{c|c|c|c}
        \toprule
        \textbf{Techniques} & \textbf{Attribution Methods} & \textbf{Local Neighborhood $\mathcal{Z}$ around $\mathbf{x}^{\{0\}}$} & \textbf{Loss Function $\ell$}\\
        \midrule
        \multirow{2}{*}{Perturbations} & Occlusion & $\mathbf{x} \odot \xi; ~\xi (\in \{0, 1\}^d) \sim \text{Random one-hot vectors}$ & Squared Error\\
        & KernelSHAP & $\mathbf{x}^{\{0\}} \odot \xi; ~\xi (\in \{0, 1\}^d) \sim \text{Shapley kernel}$ & Squared Error \\
        \midrule
        \multirow{4}{*}{Gradients} & Vanilla Gradients & $\mathbf{x} + \xi;~ \xi (\in \mathbb{R}^d) \sim \text{Normal}(0, \sigma^2), \sigma \rightarrow 0$ & Gradient Matching\\
        & Integrated Gradients & $\xi \mathbf{x}; ~\xi (\in \mathbb{R}) \sim \text{Uniform}(0,1)$ & Gradient Matching \\
        & Gradients $\times$ Input & $\xi \mathbf{x}; ~\xi (\in \mathbb{R}) \sim \text{Uniform}(a,1), a \rightarrow 1 $ & Gradient Matching\\
        & SmoothGrad & $\mathbf{x} + \xi;~ \xi (\in \mathbb{R}^d) \sim \text{Normal}(0, \sigma^2)$ & Gradient Matching  \\
        \midrule 
        \multirow{2}{*}{Linear Approximations} & LIME & $\mathbf{x} \odot \xi; ~\xi (\in \{0, 1\}^d) \sim \text{Exponential kernel}$ & Squared Error\\
        & C-LIME & $\mathbf{x} + \xi; ~\xi (\in \mathbb{R}^d) \sim \text{Normal}(0, \sigma^2)$ & Squared Error \\
        \bottomrule
    \end{tabular}
    }
    \label{tab:lfa}
\end{table}

%% file: arxiv.bbl
\begin{thebibliography}{104}
\providecommand{\natexlab}[1]{#1}
\providecommand{\url}[1]{\texttt{#1}}
\expandafter\ifx\csname urlstyle\endcsname\relax
  \providecommand{\doi}[1]{doi: #1}\else
  \providecommand{\doi}{doi: \begingroup \urlstyle{rm}\Url}\fi

\bibitem[Adebayo et~al.(2018)Adebayo, Gilmer, Muelly, Goodfellow, Hardt, and
  Kim]{adebayo2018sanity}
Julius Adebayo, Justin Gilmer, Michael Muelly, Ian Goodfellow, Moritz Hardt,
  and Been Kim.
\newblock Sanity checks for saliency maps.
\newblock \emph{Advances in neural information processing systems}, 31, 2018.

\bibitem[Adolfi et~al.(2024)Adolfi, Vilas, and
  Wareham]{adolfi2024computational}
Federico Adolfi, Martina~G Vilas, and Todd Wareham.
\newblock The computational complexity of circuit discovery for inner
  interpretability.
\newblock \emph{arXiv preprint arXiv:2410.08025}, 2024.

\bibitem[Agarwal et~al.(2022)Agarwal, Krishna, Saxena, Pawelczyk, Johnson,
  Puri, Zitnik, and Lakkaraju]{agarwal2022openxai}
Chirag Agarwal, Satyapriya Krishna, Eshika Saxena, Martin Pawelczyk, Nari
  Johnson, Isha Puri, Marinka Zitnik, and Himabindu Lakkaraju.
\newblock Openxai: Towards a transparent evaluation of model explanations.
\newblock \emph{Advances in neural information processing systems},
  35:\penalty0 15784--15799, 2022.

\bibitem[Agarwal et~al.(2021)Agarwal, Jabbari, Agarwal, Upadhyay, Wu, and
  Lakkaraju]{agarwal2021towards}
Sushant Agarwal, Shahin Jabbari, Chirag Agarwal, Sohini Upadhyay, Steven Wu,
  and Himabindu Lakkaraju.
\newblock Towards the unification and robustness of perturbation and gradient
  based explanations.
\newblock In \emph{International Conference on Machine Learning}, pages
  110--119. PMLR, 2021.

\bibitem[Ancona et~al.(2017)Ancona, Ceolini, {\"O}ztireli, and
  Gross]{ancona2017unified}
Marco Ancona, Enea Ceolini, Cengiz {\"O}ztireli, and Markus Gross.
\newblock A unified view of gradient-based attribution methods for deep neural
  networks.
\newblock \emph{arXiv preprint arXiv:1711.06104}, 2017.

\bibitem[Arrieta et~al.(2020)Arrieta, D{\'\i}az-Rodr{\'\i}guez, Del~Ser,
  Bennetot, Tabik, Barbado, Garc{\'\i}a, Gil-L{\'o}pez, Molina, Benjamins,
  et~al.]{arrieta2020explainable}
Alejandro~Barredo Arrieta, Natalia D{\'\i}az-Rodr{\'\i}guez, Javier Del~Ser,
  Adrien Bennetot, Siham Tabik, Alberto Barbado, Salvador Garc{\'\i}a, Sergio
  Gil-L{\'o}pez, Daniel Molina, Richard Benjamins, et~al.
\newblock Explainable artificial intelligence (xai): Concepts, taxonomies,
  opportunities and challenges toward responsible ai.
\newblock \emph{Information fusion}, 58:\penalty0 82--115, 2020.

\bibitem[Bach et~al.(2015)Bach, Binder, Montavon, Klauschen, M{\"u}ller, and
  Samek]{bach2015pixel}
Sebastian Bach, Alexander Binder, Gr{\'e}goire Montavon, Frederick Klauschen,
  Klaus-Robert M{\"u}ller, and Wojciech Samek.
\newblock On pixel-wise explanations for non-linear classifier decisions by
  layer-wise relevance propagation.
\newblock \emph{PloS one}, 10\penalty0 (7):\penalty0 e0130140, 2015.

\bibitem[Bae et~al.(2024)Bae, Lin, Lorraine, and Grosse]{bae2024training}
Juhan Bae, Wu~Lin, Jonathan Lorraine, and Roger~Baker Grosse.
\newblock Training data attribution via approximate unrolling.
\newblock In \emph{The Thirty-eighth Annual Conference on Neural Information
  Processing Systems}, 2024.
\newblock URL \url{https://openreview.net/forum?id=3NaqGg92KZ}.

\bibitem[Baehrens et~al.(2010)Baehrens, Schroeter, Harmeling, Kawanabe, Hansen,
  and M{\~A}{\v{z}}ller]{baehrens2010explain}
David Baehrens, Timon Schroeter, Stefan Harmeling, Motoaki Kawanabe, Katja
  Hansen, and Klaus-Robert M{\~A}{\v{z}}ller.
\newblock How to explain individual classification decisions.
\newblock \emph{Journal of Machine Learning Research}, 11\penalty0
  (Jun):\penalty0 1803--1831, 2010.

\bibitem[Barcel{\'o} et~al.(2020)Barcel{\'o}, Monet, P{\'e}rez, and
  Subercaseaux]{barcelo2020model}
Pablo Barcel{\'o}, Mika{\"e}l Monet, Jorge P{\'e}rez, and Bernardo
  Subercaseaux.
\newblock Model interpretability through the lens of computational complexity.
\newblock \emph{Advances in neural information processing systems},
  33:\penalty0 15487--15498, 2020.

\bibitem[Barshan et~al.(2020)Barshan, Brunet, and
  Dziugaite]{barshan2020relatif}
Elnaz Barshan, Marc-Etienne Brunet, and Gintare~Karolina Dziugaite.
\newblock Relatif: Identifying explanatory training samples via relative
  influence.
\newblock In \emph{International Conference on Artificial Intelligence and
  Statistics}, pages 1899--1909. PMLR, 2020.

\bibitem[Bassan et~al.(2024)Bassan, Amir, and Katz]{bassan2024local}
Shahaf Bassan, Guy Amir, and Guy Katz.
\newblock Local vs. global interpretability: A computational complexity
  perspective.
\newblock \emph{arXiv preprint arXiv:2406.02981}, 2024.

\bibitem[Bau et~al.(2020)Bau, Zhu, Strobelt, Lapedriza, Zhou, and
  Torralba]{bau2020understanding}
David Bau, Jun-Yan Zhu, Hendrik Strobelt, Agata Lapedriza, Bolei Zhou, and
  Antonio Torralba.
\newblock Understanding the role of individual units in a deep neural network.
\newblock \emph{Proceedings of the National Academy of Sciences}, 117\penalty0
  (48):\penalty0 30071--30078, 2020.

\bibitem[Bereska and Gavves(2024)]{bereska2024mechanistic}
Leonard Bereska and Efstratios Gavves.
\newblock Mechanistic interpretability for ai safety--a review.
\newblock \emph{arXiv preprint arXiv:2404.14082}, 2024.

\bibitem[Bilodeau et~al.(2024)Bilodeau, Jaques, Koh, and
  Kim]{bilodeau2024impossibility}
Blair Bilodeau, Natasha Jaques, Pang~Wei Koh, and Been Kim.
\newblock Impossibility theorems for feature attribution.
\newblock \emph{Proceedings of the National Academy of Sciences}, 121\penalty0
  (2):\penalty0 e2304406120, 2024.

\bibitem[Bordt and von Luxburg(2023)]{bordt2023shapley}
Sebastian Bordt and Ulrike von Luxburg.
\newblock From shapley values to generalized additive models and back.
\newblock In \emph{International Conference on Artificial Intelligence and
  Statistics}, pages 709--745. PMLR, 2023.

\bibitem[Bricken et~al.(2023)Bricken, Templeton, Batson, Chen, Jermyn, Conerly,
  Turner, Anil, Denison, Askell, et~al.]{bricken2023towards}
Trenton Bricken, Adly Templeton, Joshua Batson, Brian Chen, Adam Jermyn, Tom
  Conerly, Nick Turner, Cem Anil, Carson Denison, Amanda Askell, et~al.
\newblock Towards monosemanticity: Decomposing language models with dictionary
  learning.
\newblock \emph{Transformer Circuits Thread}, 2, 2023.

\bibitem[B{\"u}rger et~al.(2024)B{\"u}rger, Hamprecht, and
  Nadler]{burger2024truth}
Lennart B{\"u}rger, Fred~A Hamprecht, and Boaz Nadler.
\newblock Truth is universal: Robust detection of lies in llms.
\newblock \emph{Advances in Neural Information Processing Systems},
  37:\penalty0 138393--138431, 2024.

\bibitem[Cao et~al.(2021)Cao, Sanh, and Rush]{cao2021low}
Steven Cao, Victor Sanh, and Alexander~M Rush.
\newblock Low-complexity probing via finding subnetworks.
\newblock \emph{arXiv preprint arXiv:2104.03514}, 2021.

\bibitem[Chai et~al.(2024)Chai, Liu, Wang, Sun, Peng, and Wu]{chai2024training}
Yekun Chai, Qingyi Liu, Shuohuan Wang, Yu~Sun, Qiwei Peng, and Hua Wu.
\newblock On training data influence of gpt models.
\newblock \emph{arXiv preprint arXiv:2404.07840}, 2024.

\bibitem[Charpiat et~al.(2019)Charpiat, Girard, Felardos, and
  Tarabalka]{charpiat2019input}
Guillaume Charpiat, Nicolas Girard, Loris Felardos, and Yuliya Tarabalka.
\newblock Input similarity from the neural network perspective.
\newblock \emph{Advances in Neural Information Processing Systems}, 32, 2019.

\bibitem[Chen et~al.(2018{\natexlab{a}})Chen, Song, Wainwright, and
  Jordan]{pmlr-v80-chen18j}
Jianbo Chen, Le~Song, Martin Wainwright, and Michael Jordan.
\newblock Learning to explain: An information-theoretic perspective on model
  interpretation.
\newblock In Jennifer Dy and Andreas Krause, editors, \emph{Proceedings of the
  35th International Conference on Machine Learning}, volume~80 of
  \emph{Proceedings of Machine Learning Research}, pages 883--892,
  Stockholmsmässan, Stockholm Sweden, 10--15 Jul 2018{\natexlab{a}}. PMLR.
\newblock URL \url{http://proceedings.mlr.press/v80/chen18j.html}.

\bibitem[Chen et~al.(2018{\natexlab{b}})Chen, Song, Wainwright, and
  Jordan]{chen2018shapley}
Jianbo Chen, Le~Song, Martin~J Wainwright, and Michael~I Jordan.
\newblock L-shapley and c-shapley: Efficient model interpretation for
  structured data.
\newblock \emph{arXiv preprint arXiv:1808.02610}, 2018{\natexlab{b}}.

\bibitem[Conmy et~al.(2023)Conmy, Mavor-Parker, Lynch, Heimersheim, and
  Garriga-Alonso]{conmy2023towards}
Arthur Conmy, Augustine Mavor-Parker, Aengus Lynch, Stefan Heimersheim, and
  Adri{\`a} Garriga-Alonso.
\newblock Towards automated circuit discovery for mechanistic interpretability.
\newblock \emph{Advances in Neural Information Processing Systems},
  36:\penalty0 16318--16352, 2023.

\bibitem[Cook(1977)]{cook1977detection}
R~Dennis Cook.
\newblock Detection of influential observation in linear regression.
\newblock \emph{Technometrics}, 19\penalty0 (1):\penalty0 15--18, 1977.

\bibitem[Cook and Weisberg(1980)]{cook1980characterizations}
R~Dennis Cook and Sanford Weisberg.
\newblock Characterizations of an empirical influence function for detecting
  influential cases in regression.
\newblock \emph{Technometrics}, 22\penalty0 (4):\penalty0 495--508, 1980.

\bibitem[Cook and Weisberg(1982)]{cook1982residuals}
R~Dennis Cook and Sanford Weisberg.
\newblock Residuals and influence in regression.
\newblock \emph{NY: Chapman and Hall}, 1982.

\bibitem[Covert et~al.(2021)Covert, Lundberg, and Lee]{covert2021explaining}
Ian Covert, Scott Lundberg, and Su-In Lee.
\newblock Explaining by removing: A unified framework for model explanation.
\newblock \emph{Journal of Machine Learning Research}, 22\penalty0
  (209):\penalty0 1--90, 2021.

\bibitem[Csord{\'a}s et~al.(2020)Csord{\'a}s, van Steenkiste, and
  Schmidhuber]{csordas2020neural}
R{\'o}bert Csord{\'a}s, Sjoerd van Steenkiste, and J{\"u}rgen Schmidhuber.
\newblock Are neural nets modular? inspecting functional modularity through
  differentiable weight masks.
\newblock \emph{arXiv preprint arXiv:2010.02066}, 2020.

\bibitem[Cunningham et~al.(2023)Cunningham, Ewart, Riggs, Huben, and
  Sharkey]{cunningham2023sparse}
Hoagy Cunningham, Aidan Ewart, Logan Riggs, Robert Huben, and Lee Sharkey.
\newblock Sparse autoencoders find highly interpretable features in language
  models.
\newblock \emph{arXiv preprint arXiv:2309.08600}, 2023.

\bibitem[Dabkowski and Gal(2017)]{dabkowski2017real}
Piotr Dabkowski and Yarin Gal.
\newblock Real time image saliency for black box classifiers.
\newblock In \emph{Advances in Neural Information Processing Systems}, pages
  6970--6979, 2017.

\bibitem[De~Cao et~al.(2021)De~Cao, Aziz, and Titov]{de2021editing}
Nicola De~Cao, Wilker Aziz, and Ivan Titov.
\newblock Editing factual knowledge in language models.
\newblock \emph{arXiv preprint arXiv:2104.08164}, 2021.

\bibitem[Dhamdhere et~al.(2019)Dhamdhere, Agarwal, and
  Sundararajan]{dhamdhere2019shapley}
Kedar Dhamdhere, Ashish Agarwal, and Mukund Sundararajan.
\newblock The shapley taylor interaction index.
\newblock \emph{arXiv preprint arXiv:1902.05622}, 2019.

\bibitem[Erhan et~al.(2009)Erhan, Bengio, Courville, and
  Vincent]{erhan2009visualizing}
Dumitru Erhan, Yoshua Bengio, Aaron Courville, and Pascal Vincent.
\newblock Visualizing higher-layer features of a deep network.
\newblock \emph{University of Montreal}, 1341\penalty0 (3):\penalty0 1, 2009.

\bibitem[Fong and Vedaldi(2017)]{fong2017interpretable}
Ruth~C Fong and Andrea Vedaldi.
\newblock Interpretable explanations of black boxes by meaningful perturbation.
\newblock In \emph{Proceedings of the IEEE international conference on computer
  vision}, pages 3429--3437, 2017.

\bibitem[Geiger et~al.(2021)Geiger, Lu, Icard, and Potts]{geiger2021causal}
Atticus Geiger, Hanson Lu, Thomas Icard, and Christopher Potts.
\newblock Causal abstractions of neural networks.
\newblock \emph{Advances in Neural Information Processing Systems},
  34:\penalty0 9574--9586, 2021.

\bibitem[Geva et~al.(2023)Geva, Bastings, Filippova, and
  Globerson]{geva2023dissecting}
Mor Geva, Jasmijn Bastings, Katja Filippova, and Amir Globerson.
\newblock Dissecting recall of factual associations in auto-regressive language
  models.
\newblock \emph{arXiv preprint arXiv:2304.14767}, 2023.

\bibitem[Ghandeharioun et~al.(2024)Ghandeharioun, Caciularu, Pearce, Dixon, and
  Geva]{ghandeharioun2024patchscope}
Asma Ghandeharioun, Avi Caciularu, Adam Pearce, Lucas Dixon, and Mor Geva.
\newblock Patchscope: A unifying framework for inspecting hidden
  representations of language models.
\newblock \emph{arXiv preprint arXiv:2401.06102}, 2024.

\bibitem[Ghorbani and Zou(2019)]{ghorbani2019data}
Amirata Ghorbani and James Zou.
\newblock Data shapley: Equitable valuation of data for machine learning.
\newblock In \emph{International conference on machine learning}, pages
  2242--2251. PMLR, 2019.

\bibitem[Ghorbani and Zou(2020)]{ghorbani2020neuron}
Amirata Ghorbani and James~Y Zou.
\newblock Neuron shapley: Discovering the responsible neurons.
\newblock \emph{Advances in neural information processing systems},
  33:\penalty0 5922--5932, 2020.

\bibitem[Grosse et~al.(2023)Grosse, Bae, Anil, Elhage, Tamkin, Tajdini,
  Steiner, Li, Durmus, Perez, et~al.]{grosse2023studying}
Roger Grosse, Juhan Bae, Cem Anil, Nelson Elhage, Alex Tamkin, Amirhossein
  Tajdini, Benoit Steiner, Dustin Li, Esin Durmus, Ethan Perez, et~al.
\newblock Studying large language model generalization with influence
  functions.
\newblock \emph{arXiv preprint arXiv:2308.03296}, 2023.

\bibitem[Guidotti et~al.(2018)Guidotti, Monreale, Turini, Pedreschi, and
  Giannotti]{guidotti2018survey}
Riccardo Guidotti, Anna Monreale, Franco Turini, Dino Pedreschi, and Fosca
  Giannotti.
\newblock A survey of methods for explaining black box models.
\newblock \emph{arXiv preprint arXiv:1802.01933}, 2018.

\bibitem[Guo et~al.(2021)Guo, Rajani, Hase, Bansal, and
  Xiong]{guo-etal-2021-fastif}
Han Guo, Nazneen Rajani, Peter Hase, Mohit Bansal, and Caiming Xiong.
\newblock {F}ast{IF}: Scalable influence functions for efficient model
  interpretation and debugging.
\newblock In \emph{Proceedings of the 2021 Conference on Empirical Methods in
  Natural Language Processing}. Association for Computational Linguistics,
  2021.
\newblock \doi{10.18653/v1/2021.emnlp-main.808}.

\bibitem[Hammoudeh and Lowd(2024)]{hammoudeh2024training}
Zayd Hammoudeh and Daniel Lowd.
\newblock Training data influence analysis and estimation: A survey.
\newblock \emph{Machine Learning}, 113\penalty0 (5):\penalty0 2351--2403, 2024.

\bibitem[Han et~al.(2022)Han, Srinivas, and Lakkaraju]{lfa2022}
Tessa Han, Suraj Srinivas, and Himabindu Lakkaraju.
\newblock Which explanation should i choose? a function approximation
  perspective to characterizing post hoc explanations.
\newblock In \emph{Advances in Neural Information Processing Systems
  (NeurIPS)}, 2022.

\bibitem[Hara et~al.(2019)Hara, Nitanda, and Maehara]{hara2019data}
Satoshi Hara, Atsushi Nitanda, and Takanori Maehara.
\newblock Data cleansing for models trained with sgd.
\newblock \emph{Advances in Neural Information Processing Systems}, 32, 2019.

\bibitem[Hase et~al.(2024)Hase, Bansal, Kim, and Ghandeharioun]{hase2024does}
Peter Hase, Mohit Bansal, Been Kim, and Asma Ghandeharioun.
\newblock Does localization inform editing? surprising differences in
  causality-based localization vs. knowledge editing in language models.
\newblock \emph{Advances in Neural Information Processing Systems}, 36, 2024.

\bibitem[Horel and Giesecke(2020)]{horel2020significance}
Enguerrand Horel and Kay Giesecke.
\newblock Significance tests for neural networks.
\newblock \emph{Journal of Machine Learning Research}, 21\penalty0
  (227):\penalty0 1--29, 2020.

\bibitem[Horel and Giesecke(2022)]{horel2022computationally}
Enguerrand Horel and Kay Giesecke.
\newblock Computationally efficient feature significance and importance for
  predictive models.
\newblock In \emph{Proceedings of the Third ACM International Conference on AI
  in Finance}, pages 300--307, 2022.

\bibitem[Ilyas et~al.(2022)Ilyas, Park, Engstrom, Leclerc, and
  Madry]{ilyas2022datamodels}
Andrew Ilyas, Sung~Min Park, Logan Engstrom, Guillaume Leclerc, and Aleksander
  Madry.
\newblock Datamodels: Predicting predictions from training data.
\newblock \emph{arXiv preprint arXiv:2202.00622}, 2022.

\bibitem[Janizek et~al.(2021)Janizek, Sturmfels, and
  Lee]{janizek2021explaining}
Joseph~D Janizek, Pascal Sturmfels, and Su-In Lee.
\newblock Explaining explanations: Axiomatic feature interactions for deep
  networks.
\newblock \emph{Journal of Machine Learning Research}, 22\penalty0
  (104):\penalty0 1--54, 2021.

\bibitem[Jeyakumar et~al.(2020)Jeyakumar, Noor, Cheng, Garcia, and
  Srivastava]{jeyakumar2020can}
Jeya~Vikranth Jeyakumar, Joseph Noor, Yu-Hsi Cheng, Luis Garcia, and Mani
  Srivastava.
\newblock How can i explain this to you? an empirical study of deep neural
  network explanation methods.
\newblock \emph{Advances in neural information processing systems},
  33:\penalty0 4211--4222, 2020.

\bibitem[Jia et~al.(2019)Jia, Dao, Wang, Hubis, Gurel, Li, Zhang, Spanos, and
  Song]{jia2019efficient}
Ruoxi Jia, David Dao, Boxin Wang, Frances~Ann Hubis, Nezihe~Merve Gurel, Bo~Li,
  Ce~Zhang, Costas~J Spanos, and Dawn Song.
\newblock Efficient task-specific data valuation for nearest neighbor
  algorithms.
\newblock \emph{arXiv preprint arXiv:1908.08619}, 2019.

\bibitem[Jia et~al.(2021)Jia, Wu, Sun, Xu, Dao, Kailkhura, Zhang, Li, and
  Song]{jia2021scalability}
Ruoxi Jia, Fan Wu, Xuehui Sun, Jiacen Xu, David Dao, Bhavya Kailkhura,
  Ce~Zhang, Bo~Li, and Dawn Song.
\newblock Scalability vs. utility: Do we have to sacrifice one for the other in
  data importance quantification?
\newblock In \emph{Proceedings of the IEEE/CVF Conference on Computer Vision
  and Pattern Recognition}, pages 8239--8247, 2021.

\bibitem[Koh and Liang(2017)]{koh2017understanding}
Pang~Wei Koh and Percy Liang.
\newblock Understanding black-box predictions via influence functions.
\newblock In \emph{International conference on machine learning}, pages
  1885--1894. PMLR, 2017.

\bibitem[Koh et~al.(2020)Koh, Nguyen, Tang, Mussmann, Pierson, Kim, and
  Liang]{koh2020concept}
Pang~Wei Koh, Thao Nguyen, Yew~Siang Tang, Stephen Mussmann, Emma Pierson, Been
  Kim, and Percy Liang.
\newblock Concept bottleneck models.
\newblock In \emph{International conference on machine learning}, pages
  5338--5348. PMLR, 2020.

\bibitem[Krishna* et~al.(2024)Krishna*, Han*, Gu, Wu, Jabbari, and
  Lakkaraju]{disagreement2024}
Satyapriya Krishna*, Tessa Han*, Alex Gu, Steven Wu, Shahin Jabbari, and
  Himabindu Lakkaraju.
\newblock The disagreement problem in explainable machine learning: A
  practitioner's perspective.
\newblock \emph{Transactions on Machine Learning Research (TMLR)}, 2024.

\bibitem[Kwon and Zou(2022)]{kwon2022beta}
Yongchan Kwon and James Zou.
\newblock Beta shapley: a unified and noise-reduced data valuation framework
  for machine learning.
\newblock In \emph{International Conference on Artificial Intelligence and
  Statistics}, pages 8780--8802. PMLR, 2022.

\bibitem[Li and Janson(2024)]{li2024optimal}
Maximilian Li and Lucas Janson.
\newblock Optimal ablation for interpretability.
\newblock \emph{arXiv preprint arXiv:2409.09951}, 2024.

\bibitem[Lin et~al.(2022)Lin, Zhang, L{\'e}cuyer, Li, Panda, and
  Sen]{lin2022measuring}
Jinkun Lin, Anqi Zhang, Mathias L{\'e}cuyer, Jinyang Li, Aurojit Panda, and
  Siddhartha Sen.
\newblock Measuring the effect of training data on deep learning predictions
  via randomized experiments.
\newblock In \emph{International Conference on Machine Learning}, pages
  13468--13504. PMLR, 2022.

\bibitem[Lundberg and Lee(2017)]{lundberg2017unified}
Scott~M Lundberg and Su-In Lee.
\newblock A unified approach to interpreting model predictions.
\newblock In \emph{Advances in Neural Information Processing Systems}, pages
  4768--4777, 2017.

\bibitem[Marks and Tegmark(2023)]{marks2023geometry}
Samuel Marks and Max Tegmark.
\newblock The geometry of truth: Emergent linear structure in large language
  model representations of true/false datasets.
\newblock \emph{arXiv preprint arXiv:2310.06824}, 2023.

\bibitem[Marks et~al.(2024)Marks, Rager, Michaud, Belinkov, Bau, and
  Mueller]{marks2024sparse}
Samuel Marks, Can Rager, Eric~J Michaud, Yonatan Belinkov, David Bau, and Aaron
  Mueller.
\newblock Sparse feature circuits: Discovering and editing interpretable causal
  graphs in language models.
\newblock \emph{arXiv preprint arXiv:2403.19647}, 2024.

\bibitem[Meng et~al.(2022)Meng, Bau, Andonian, and Belinkov]{meng2022locating}
Kevin Meng, David Bau, Alex Andonian, and Yonatan Belinkov.
\newblock Locating and editing factual associations in gpt.
\newblock \emph{Advances in Neural Information Processing Systems},
  35:\penalty0 17359--17372, 2022.

\bibitem[Mirzasoleiman et~al.(2020)Mirzasoleiman, Bilmes, and
  Leskovec]{mirzasoleiman2020coresets}
Baharan Mirzasoleiman, Jeff Bilmes, and Jure Leskovec.
\newblock Coresets for data-efficient training of machine learning models.
\newblock In \emph{International Conference on Machine Learning}, pages
  6950--6960. PMLR, 2020.

\bibitem[Mitchell et~al.(2021)Mitchell, Lin, Bosselut, Finn, and
  Manning]{mitchell2021fast}
Eric Mitchell, Charles Lin, Antoine Bosselut, Chelsea Finn, and Christopher~D
  Manning.
\newblock Fast model editing at scale.
\newblock \emph{arXiv preprint arXiv:2110.11309}, 2021.

\bibitem[Mueller et~al.(2024)Mueller, Brinkmann, Li, Marks, Pal, Prakash,
  Rager, Sankaranarayanan, Sharma, Sun, et~al.]{mueller2024quest}
Aaron Mueller, Jannik Brinkmann, Millicent Li, Samuel Marks, Koyena Pal, Nikhil
  Prakash, Can Rager, Aruna Sankaranarayanan, Arnab~Sen Sharma, Jiuding Sun,
  et~al.
\newblock The quest for the right mediator: A history, survey, and theoretical
  grounding of causal interpretability.
\newblock \emph{arXiv preprint arXiv:2408.01416}, 2024.

\bibitem[Nanda(2023)]{nanda2023attribution}
Neel Nanda.
\newblock Attribution patching: Activation patching at industrial scale.
\newblock \emph{URL: https://www. neelnanda.
  io/mechanistic-interpretability/attribution-patching}, 2023.

\bibitem[Oesterling et~al.(2024)Oesterling, Bhalla, Venkatasubramanian, and
  Lakkaraju]{oesterling2024operationalizing}
Alex Oesterling, Usha Bhalla, Suresh Venkatasubramanian, and Himabindu
  Lakkaraju.
\newblock Operationalizing the blueprint for an ai bill of rights:
  Recommendations for practitioners, researchers, and policy makers.
\newblock \emph{arXiv preprint arXiv:2407.08689}, 2024.

\bibitem[Olsson et~al.(2022)Olsson, Elhage, Nanda, Joseph, DasSarma, Henighan,
  Mann, Askell, Bai, Chen, et~al.]{olsson2022context}
Catherine Olsson, Nelson Elhage, Neel Nanda, Nicholas Joseph, Nova DasSarma,
  Tom Henighan, Ben Mann, Amanda Askell, Yuntao Bai, Anna Chen, et~al.
\newblock In-context learning and induction heads.
\newblock \emph{arXiv preprint arXiv:2209.11895}, 2022.

\bibitem[Park et~al.(2023)Park, Georgiev, Ilyas, Leclerc, and
  Madry]{park2023trak}
Sung~Min Park, Kristian Georgiev, Andrew Ilyas, Guillaume Leclerc, and
  Aleksander Madry.
\newblock Trak: Attributing model behavior at scale.
\newblock \emph{arXiv preprint arXiv:2303.14186}, 2023.

\bibitem[Patel et~al.(2021)Patel, Strobel, and Zick]{patel2021high}
Neel Patel, Martin Strobel, and Yair Zick.
\newblock High dimensional model explanations: An axiomatic approach.
\newblock In \emph{Proceedings of the 2021 ACM Conference on Fairness,
  Accountability, and Transparency}, pages 401--411, 2021.

\bibitem[Pearl(2022)]{pearl2022direct}
Judea Pearl.
\newblock Direct and indirect effects.
\newblock In \emph{Probabilistic and causal inference: the works of Judea
  Pearl}, pages 373--392. Association for Computing Machinery and Morgan \&
  Claypool Publishers, 2022.

\bibitem[Petsiuk(2018)]{petsiuk2018rise}
V~Petsiuk.
\newblock Rise: Randomized input sampling for explanation of black-box models.
\newblock \emph{arXiv preprint arXiv:1806.07421}, 2018.

\bibitem[Pruthi et~al.(2020)Pruthi, Liu, Kale, and
  Sundararajan]{pruthi2020estimating}
Garima Pruthi, Frederick Liu, Satyen Kale, and Mukund Sundararajan.
\newblock Estimating training data influence by tracing gradient descent.
\newblock \emph{Advances in Neural Information Processing Systems},
  33:\penalty0 19920--19930, 2020.

\bibitem[Rai et~al.(2024)Rai, Zhou, Feng, Saparov, and Yao]{rai2024practical}
Daking Rai, Yilun Zhou, Shi Feng, Abulhair Saparov, and Ziyu Yao.
\newblock A practical review of mechanistic interpretability for
  transformer-based language models.
\newblock \emph{arXiv preprint arXiv:2407.02646}, 2024.

\bibitem[R{\"a}uker et~al.(2023)R{\"a}uker, Ho, Casper, and
  Hadfield-Menell]{rauker2023toward}
Tilman R{\"a}uker, Anson Ho, Stephen Casper, and Dylan Hadfield-Menell.
\newblock Toward transparent ai: A survey on interpreting the inner structures
  of deep neural networks.
\newblock In \emph{2023 ieee conference on secure and trustworthy machine
  learning (satml)}, pages 464--483. IEEE, 2023.

\bibitem[Ribeiro et~al.(2016)Ribeiro, Singh, and Guestrin]{ribeiro2016should}
Marco~Tulio Ribeiro, Sameer Singh, and Carlos Guestrin.
\newblock Why should i trust you?: Explaining the predictions of any
  classifier.
\newblock In \emph{Proceedings of the 22nd ACM SIGKDD International Conference
  on Knowledge Discovery and Data Mining}, pages 1135--1144. ACM, 2016.

\bibitem[Ronneberger et~al.(2015)Ronneberger, Fischer, and
  Brox]{ronneberger2015u}
Olaf Ronneberger, Philipp Fischer, and Thomas Brox.
\newblock U-net: Convolutional networks for biomedical image segmentation.
\newblock In \emph{Medical image computing and computer-assisted
  intervention--MICCAI 2015: 18th international conference, Munich, Germany,
  October 5-9, 2015, proceedings, part III 18}, pages 234--241. Springer, 2015.

\bibitem[Saphra and Wiegreffe(2024)]{saphra2024mechanistic}
Naomi Saphra and Sarah Wiegreffe.
\newblock Mechanistic?
\newblock \emph{arXiv preprint arXiv:2410.09087}, 2024.

\bibitem[Schioppa et~al.(2022)Schioppa, Zablotskaia, Vilar, and
  Sokolov]{schioppa2022scaling}
Andrea Schioppa, Polina Zablotskaia, David Vilar, and Artem Sokolov.
\newblock Scaling up influence functions.
\newblock In \emph{Proceedings of the AAAI Conference on Artificial
  Intelligence}, volume~36, pages 8179--8186, 2022.

\bibitem[Selvaraju et~al.(2016)Selvaraju, Das, Vedantam, Cogswell, Parikh, and
  Batra]{selvaraju2016grad}
Ramprasaath~R Selvaraju, Abhishek Das, Ramakrishna Vedantam, Michael Cogswell,
  Devi Parikh, and Dhruv Batra.
\newblock Grad-cam: Why did you say that?
\newblock \emph{arXiv preprint arXiv:1611.07450}, 2016.

\bibitem[Shah et~al.(2024)Shah, Ilyas, and Madry]{shah2024decomposing}
Harshay Shah, Andrew Ilyas, and Aleksander Madry.
\newblock Decomposing and editing predictions by modeling model computation.
\newblock \emph{arXiv preprint arXiv:2404.11534}, 2024.

\bibitem[Shapley(1953)]{shapley1953value}
Lloyd~S Shapley.
\newblock A value for n-person games.
\newblock \emph{Contributions to the Theory of Games}, 2, 1953.

\bibitem[Shrikumar et~al.(2017)Shrikumar, Greenside, and
  Kundaje]{shrikumar2017learning}
Avanti Shrikumar, Peyton Greenside, and Anshul Kundaje.
\newblock Learning important features through propagating activation
  differences.
\newblock In \emph{International conference on machine learning}, pages
  3145--3153. PMlR, 2017.

\bibitem[Simonyan et~al.(2013)Simonyan, Vedaldi, and
  Zisserman]{simonyan2013deep}
Karen Simonyan, Andrea Vedaldi, and Andrew Zisserman.
\newblock Deep inside convolutional networks: Visualising image classification
  models and saliency maps.
\newblock \emph{arXiv preprint arXiv:1312.6034}, 2013.

\bibitem[Smilkov et~al.(2017)Smilkov, Thorat, Kim, Vi{\'e}gas, and
  Wattenberg]{smilkov2017smoothgrad}
Daniel Smilkov, Nikhil Thorat, Been Kim, Fernanda Vi{\'e}gas, and Martin
  Wattenberg.
\newblock Smoothgrad: removing noise by adding noise.
\newblock \emph{arXiv preprint arXiv:1706.03825}, 2017.

\bibitem[Sokol and Flach(2020)]{sokol2020limetree}
Kacper Sokol and Peter Flach.
\newblock Limetree: Interactively customisable explanations based on local
  surrogate multi-output regression trees.
\newblock \emph{arXiv}, 2020.

\bibitem[Springenberg et~al.(2014)Springenberg, Dosovitskiy, Brox, and
  Riedmiller]{springenberg2014striving}
Jost~Tobias Springenberg, Alexey Dosovitskiy, Thomas Brox, and Martin
  Riedmiller.
\newblock Striving for simplicity: The all convolutional net.
\newblock \emph{arXiv preprint arXiv:1412.6806}, 2014.

\bibitem[Sun et~al.(2023)Sun, Liu, Bair, and Kolter]{sun2023simple}
Mingjie Sun, Zhuang Liu, Anna Bair, and J~Zico Kolter.
\newblock A simple and effective pruning approach for large language models.
\newblock \emph{arXiv preprint arXiv:2306.11695}, 2023.

\bibitem[Sundararajan and Najmi(2020)]{sundararajan2020many}
Mukund Sundararajan and Amir Najmi.
\newblock The many shapley values for model explanation.
\newblock In \emph{International conference on machine learning}, pages
  9269--9278. PMLR, 2020.

\bibitem[Sundararajan et~al.(2017)Sundararajan, Taly, and
  Yan]{sundararajan2017axiomatic}
Mukund Sundararajan, Ankur Taly, and Qiqi Yan.
\newblock Axiomatic attribution for deep networks.
\newblock \emph{arXiv preprint arXiv:1703.01365}, 2017.

\bibitem[Syed et~al.(2023)Syed, Rager, and Conmy]{syed2023attribution}
Aaquib Syed, Can Rager, and Arthur Conmy.
\newblock Attribution patching outperforms automated circuit discovery.
\newblock \emph{arXiv preprint arXiv:2310.10348}, 2023.

\bibitem[Tukey(1958)]{tukey1958bias}
John Tukey.
\newblock Bias and confidence in not quite large samples.
\newblock \emph{Ann. Math. Statist.}, 29:\penalty0 614, 1958.

\bibitem[Vig et~al.(2020)Vig, Gehrmann, Belinkov, Qian, Nevo, Singer, and
  Shieber]{vig2020investigating}
Jesse Vig, Sebastian Gehrmann, Yonatan Belinkov, Sharon Qian, Daniel Nevo,
  Yaron Singer, and Stuart Shieber.
\newblock Investigating gender bias in language models using causal mediation
  analysis.
\newblock \emph{Advances in neural information processing systems},
  33:\penalty0 12388--12401, 2020.

\bibitem[Vilas et~al.(2024)Vilas, Adolfi, Poeppel, and Roig]{vilas2024position}
Martina~G Vilas, Federico Adolfi, David Poeppel, and Gemma Roig.
\newblock Position: An inner interpretability framework for ai inspired by
  lessons from cognitive neuroscience.
\newblock \emph{arXiv preprint arXiv:2406.01352}, 2024.

\bibitem[Wang and Jia(2023)]{wang2023data}
Jiachen~T Wang and Ruoxi Jia.
\newblock Data banzhaf: A robust data valuation framework for machine learning.
\newblock In \emph{International Conference on Artificial Intelligence and
  Statistics}, pages 6388--6421. PMLR, 2023.

\bibitem[Wang et~al.(2022)Wang, Variengien, Conmy, Shlegeris, and
  Steinhardt]{wang2022interpretability}
Kevin Wang, Alexandre Variengien, Arthur Conmy, Buck Shlegeris, and Jacob
  Steinhardt.
\newblock Interpretability in the wild: a circuit for indirect object
  identification in gpt-2 small.
\newblock \emph{arXiv preprint arXiv:2211.00593}, 2022.

\bibitem[Wang et~al.(2024)Wang, Zhang, Guo, and Shen]{wang2024gradient}
Yongjie Wang, Tong Zhang, Xu~Guo, and Zhiqi Shen.
\newblock Gradient based feature attribution in explainable ai: A technical
  review.
\newblock \emph{arXiv preprint arXiv:2403.10415}, 2024.

\bibitem[Yan and Procaccia(2021)]{yan2021if}
Tom Yan and Ariel~D Procaccia.
\newblock If you like shapley then you’ll love the core.
\newblock In \emph{Proceedings of the AAAI Conference on Artificial
  Intelligence}, volume~35, pages 5751--5759, 2021.

\bibitem[Yoon et~al.(2020)Yoon, Arik, and Pfister]{yoon2020data}
Jinsung Yoon, Sercan Arik, and Tomas Pfister.
\newblock Data valuation using reinforcement learning.
\newblock In \emph{International Conference on Machine Learning}, pages
  10842--10851. PMLR, 2020.

\bibitem[Zeiler and Fergus(2014)]{zeiler2014visualizing}
Matthew~D Zeiler and Rob Fergus.
\newblock Visualizing and understanding convolutional networks.
\newblock In \emph{Computer Vision--ECCV 2014: 13th European Conference,
  Zurich, Switzerland, September 6-12, 2014, Proceedings, Part I 13}, pages
  818--833. Springer, 2014.

\bibitem[Zhang et~al.(2022)Zhang, Liu, Shah, and Sun]{zhang2022gstarx}
Shichang Zhang, Yozen Liu, Neil Shah, and Yizhou Sun.
\newblock Gstarx: Explaining graph neural networks with structure-aware
  cooperative games.
\newblock \emph{Advances in Neural Information Processing Systems},
  35:\penalty0 19810--19823, 2022.

\bibitem[Zou et~al.(2023)Zou, Phan, Chen, Campbell, Guo, Ren, Pan, Yin,
  Mazeika, Dombrowski, et~al.]{zou2023representation}
Andy Zou, Long Phan, Sarah Chen, James Campbell, Phillip Guo, Richard Ren,
  Alexander Pan, Xuwang Yin, Mantas Mazeika, Ann-Kathrin Dombrowski, et~al.
\newblock Representation engineering: A top-down approach to ai transparency.
\newblock \emph{arXiv preprint arXiv:2310.01405}, 2023.

\end{thebibliography}
